\documentclass[lettersize,journal]{IEEEtran}
\usepackage{amsmath,amsfonts}
\usepackage{algorithmic}
\usepackage{algorithm}
\usepackage{array}
\usepackage[caption=false,font=normalsize,labelfont=sf,textfont=sf]{subfig}
\usepackage{textcomp}
\usepackage{stfloats}
\usepackage{url}
\usepackage{verbatim}
\usepackage{graphicx}
\usepackage{cite}
\usepackage{bm}
\usepackage{multirow}
\usepackage{booktabs}
\usepackage{colortbl}
\usepackage{xcolor}
\usepackage{amssymb}

\begin{document}

\title{The Pictorial Cortex: Zero-Shot Cross-Subject fMRI-to-Image Reconstruction via Compositional Latent Modeling}


\author{Jingyang Huo,~Yikai Wang,~Yanwei Fu,~Jianfeng Feng 
\thanks{Jingyang Huo and Jianfeng Feng are with the Institute of Science and Technology for Brain-inspired Intelligence (ISTBI), Fudan University. 
Jianfeng Fen, and Yanwei Fu are also with the School of Data Science, Fudan University, Shanghai 200437, China,  and Fudan ISTBI–ZJNU Algorithm Centre for Brain-Inspired Intelligence, Zhejiang Normal University, Jinhua 321017, China. Yanwei Fu is also with Shanghai Innovation Institute, Shanghai 200080, China (e-mail: jyhuo22@m.fudan.edu.cn, jffeng@fudan.edu.cn, yanweifu@fudan.edu.cn).
Yanwei Fu is the corresponding author.
}% <-this % stops a space
\thanks{Yikai Wang is with Nanyang Technological University (e-mail: yi-kai.wang@outlook.com).
}
\thanks{Manuscript received April 19, 2021; revised August 16, 2021.}
}

\markboth{Journal of \LaTeX\ Class Files,~Vol.~14, No.~8, August~2021}%
{Shell \MakeLowercase{\textit{et al.}}: A Sample Article Using IEEEtran.cls for IEEE Journals}


\maketitle

\begin{abstract} 
Decoding visual experiences from human brain activity remains a central challenge at the intersection of neuroscience, neuroimaging, and artificial intelligence. A critical obstacle is the inherent variability of cortical responses: neural activity elicited by the same visual stimulus differs across individuals and trials due to anatomical, functional, cognitive, and experimental factors, making fMRI-to-image reconstruction highly non-injective. In this paper, we tackle a challenging yet practically meaningful problem: zero-shot cross-subject fMRI-to-image reconstruction (ZS-CS fMRI2Image), where the visual experience of a previously unseen individual must be reconstructed without subject-specific training. 
To enable principled evaluation, we present a unified cortical-surface  dataset -- \textbf{UniCortex-fMRI}, assembled from multiple visual-stimulus fMRI datasets to provide broad coverage of subjects and stimuli. Our UniCortex-fMRI is particularly processed by standardized data formats to make it possible to explore this possibility in the zero-shot scenario of cross-subject fMRI-to-image reconstruction. 
To tackle the modeling challenge, we propose \textbf{the Pictorial Cortex (PictorialCortex)}, which models fMRI activity using a \textit{compositional latent formulation} that structures stimulus-driven representations under subject-, dataset-, and trial-related variability. PictorialCortex operates in a universal cortical latent space and implements this formulation through a latent factorization--composition module, reinforced by paired factorization and re-factorizing consistency regularization. During inference, surrogate latents synthesized under multiple seen-subject conditions are aggregated to guide diffusion-based image synthesis for unseen subjects.
Extensive experiments demonstrate that PictorialCortex substantially improves zero-shot cross-subject visual reconstruction, highlighting the benefits of compositional latent modeling and multi-dataset training. The dataset, code, and models will be released and shared to the community.

\end{abstract}


\section{Introduction}

\IEEEPARstart{D}{ecoding} visual experiences from human brain activity remains a central challenge in neuroscience and a rapidly advancing frontier bridging neuroimaging, computer vision, and artificial intelligence.
In recent years, this pursuit has increasingly focused on a concrete and challenging task: reconstructing visual images directly from functional magnetic resonance imaging (fMRI) signals~\cite{glover2011overview}.
At its core, fMRI-to-image reconstruction seeks to answer a fundamental question:
\emph{can the visual content perceived by a human observer be reliably inferred from their brain activity alone?}
In  a typical stimulus-driven paradigm, subjects view visual stimuli while large-scale, distributed neural responses are elicited across the cortex and captured by fMRI.
The objective of fMRI-to-image reconstruction is to transform these indirect and noisy neural measurements into images that are consistent with the observer’s perceptual experience.

\begin{figure}[t]
\centering
\includegraphics[width=\linewidth]{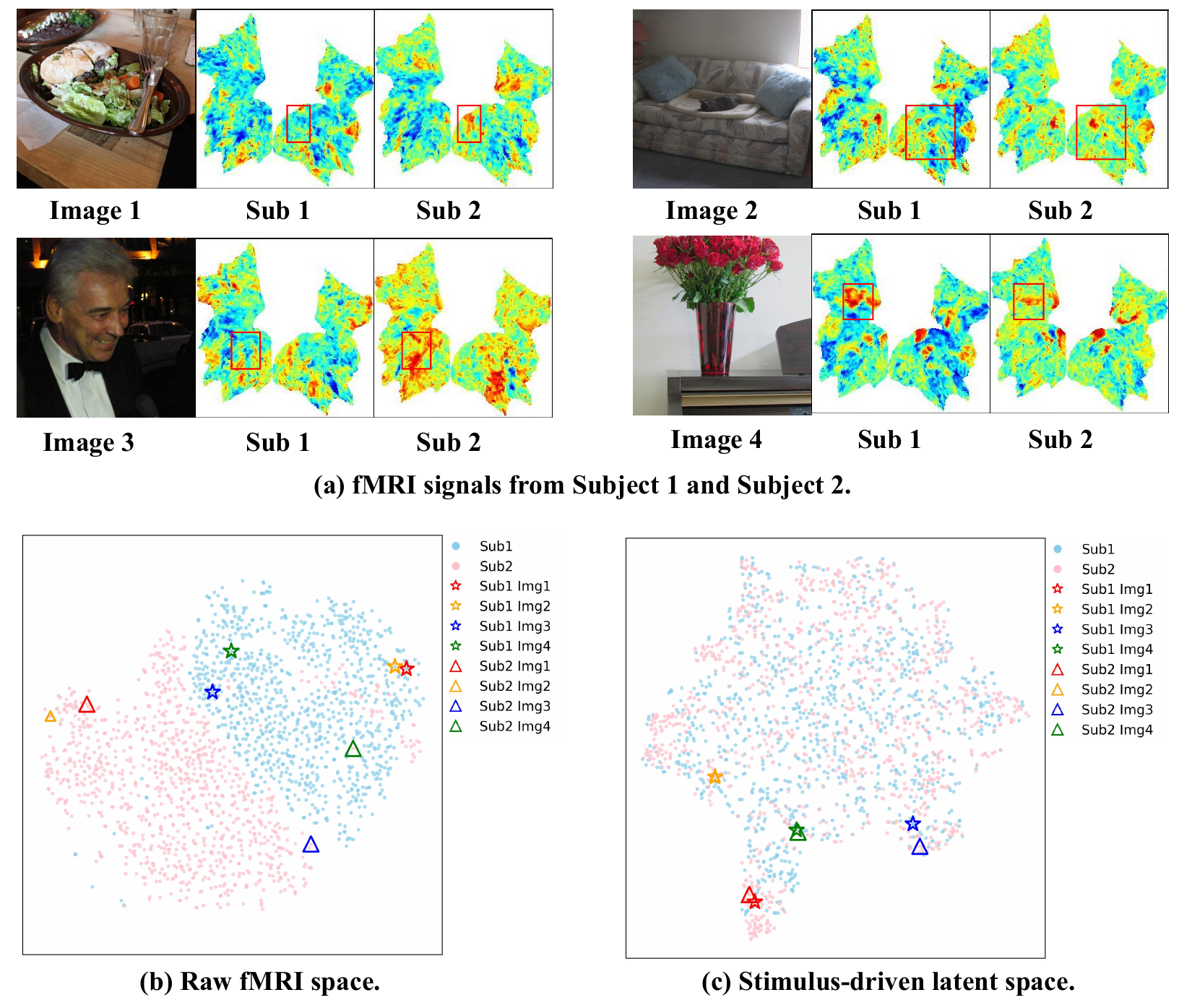}
\caption{
fMRI responses of Subject 1 and Subject 2 under the same images: 
(a) raw fMRI responses on the cortical surface, 
(b) t-SNE of raw fMRI space, and (c) t-SNE of PictorialCortex stimulus-driven latent space. Colors indicate different images; star markers denote Subject 1, and upward triangles denote Subject 2. In the raw fMRI space, responses cluster by subject rather than image, while in the stimulus-driven latent space, the same images from different subjects are closely aligned. 
}
\label{fig:fmri_diff}
\end{figure}

A central challenge of fMRI-to-image reconstruction stems from the inherent subjectivity of cortical responses.
Human perception is not a direct replication of the external world, but rather the result of complex interactions between objective sensory input and subjective interpretation.
Consequently, the fMRI responses elicited by the same visual stimulus can differ substantially across individuals, and even across repeated presentations within the same individual~\cite{aguirre1998variability}. 
This subjectivity manifests at multiple levels. 
(i) \emph{Inter-subject variability} arises from differences in cortical anatomy, functional topography, vascular coupling, and baseline neural dynamics, leading different individuals to exhibit distinct fMRI activation patterns under identical visual input~\cite{mueller2013individual,laumann2015functional,guntupalli2016model}.
As shown in Figure~\ref{fig:fmri_diff}(a), the same image evokes markedly different cortical activation patterns across subjects.
When visualized in the raw fMRI latent space (Figure~\ref{fig:fmri_diff}(b)), responses are dominated by subject identity rather than stimulus content: fMRI patterns from the same subject are closer to each other even when the viewed images differ, while responses to the same image from different subjects are widely separated.
This subject-dominated organization indicates individual cortical characteristics can overshadow objective image responses in fMRI signals.
(ii) In addition to  inter-subject variability, substantial \emph{intra-subject (trial-wise) variability} further complicates decoding.
Repeated presentations of the same stimulus can elicit different fMRI responses due to fluctuations in attention, cognitive state, neural adaptation, slow baseline drifts, and measurement noise~\cite{aguirre1998variability,poldrack2015long,summerfield2009expectation}.
These sources of variability also interact with heterogeneity in experimental protocols, including stimulus design, scanning parameters, and preprocessing pipelines, introducing additional dataset-dependent biases~\cite{yu2018statistical}. 
Together, these factors render the mapping from visual stimuli to fMRI responses highly non-injective, in which multiple, subject- and context-dependent neural patterns correspond to the same perceptual content. This non-injectivity poses a fundamental obstacle to robust and generalizable fMRI-to-image reconstruction.

In this paper, we address a substantially more challenging yet practically and scientifically meaningful objective: zero-shot cross-subject fMRI-to-image reconstruction, abbreviated as ZS-CS fMRI2Image. In this setting, the goal is to reconstruct the visual experience of a previously unseen individual without any subject-specific training. This formulation more accurately reflects real-world neuroimaging constraints, where collecting extensive fMRI data for each new subject is costly, time-consuming, and often impractical. Achieving ZS-CS fMRI2Image requires learning representations that are both robust and transferable across individuals, which in turn demands explicitly disentangling stimulus-driven visual content from the heterogeneous subject-specific, dataset-dependent, and trial-wise factors entangled in fMRI signals. To the best of our knowledge, zero-shot cross-subject fMRI-to-image reconstruction has received little \textit{systematic investigation}, primarily due to the intertwined model and dataset challenges inherent to this task.

\noindent \textbf{Model Challenge}. Recent advances in generative modeling, particularly diffusion-based image synthesis, have substantially revitalized research on fMRI-to-image reconstruction. 
Our prior work, \textit{NeuroPictor}~\cite{neuropictor}, introduced a diffusion-based framework that directly modulates image generation using multi-level conditioning extracted from fMRI signals. By mapping fMRI data onto a unified cortical-surface representation, NeuroPictor showed that multi-individual pretraining can improve reconstruction quality for single-subject fine-tuning. It achieves precise control over both semantic and spatial aspects of reconstructed images, producing high-fidelity results. Nevertheless, 
our NeuroPictor, like most existing approaches~\cite{neuropictor, scotti2024mindeye2, chen2022seeing, scotti2023reconstructing, ozcelik2023natural}, operates predominantly in a within-subject paradigm, in which decoders are trained (or fine-tuned) and evaluated on the same individual. While effective under controlled experimental settings, this paradigm implicitly treats subject- and trial-specific variability as noise rather than explicitly modeling it. As a result, these methods typically encode fMRI signals into a single visual latent space, collapsing heterogeneous sources of variability into a monolithic representation. When combined with the high dimensionality of fMRI data and pronounced inter-subject heterogeneity, this design encourages overfitting to idiosyncratic subject–stimulus associations, thereby harming robustness and severely limiting generalization to unseen individuals and novel stimuli. Fundamentally, this limitation arises from inadequate modeling of  multiple factors underlying cortical responses, as fMRI activity reflects a complex mixture of stimulus-driven visual information, individual anatomical and functional characteristics, cognitive state, and experimental protocol.

\noindent \textbf{Dataset Challenge}. Progress in zero-shot cross-subject fMRI-to-image reconstruction has also been constrained by the absence of an appropriate dataset-level testbed.
Existing fMRI datasets exhibit a pronounced imbalance between subject diversity and stimulus variability, limiting principled evaluation of cross-subject generalization. Large-scale visual fMRI datasets such as NSD~\cite{allen2022massive}, BOLD5000~\cite{chang2019bold5000}, and NOD~\cite{nod2023} provide thousands of image–fMRI pairs per subject, but include only a small number of individuals, encouraging models to overfit subject-specific regularities rather than learn subject-agnostic representations. In contrast, datasets such as HCP-Movie~\cite{hcp2013wu} offer substantially broader subject coverage but extremely sparse stimulus exposure per subject, making it difficult to disentangle stimulus-driven visual information from noise. As a result, neither dimension is sufficiently represented in isolation, impeding rigorous study of zero-shot cross-subject fMRI-to-image reconstruction.
To overcome this limitation, we present UniCortex-fMRI, a unified cortical-surface fMRI dataset specifically designed as a principled testbed for cross-subject generalization. UniCortex-fMRI integrates four heterogeneous visual-stimulus datasets—NSD, BOLD5000, NOD, and HCP-Movie—into a common fsLR cortical surface representation with standardized preprocessing and unified data formats, making the dataset directly and publicly usable by the research community. As a very large unified cortical-surface fMRI dataset for this task, UniCortex-fMRI provides extensive coverage of both subjects and stimuli and supports a principled evaluation protocol with controlled seen/unseen subject splits and consistent task definitions, enabling robust and systematic assessment of zero-shot cross-subject reconstruction.

\noindent \textbf{Proposed Framework}. We further address the model challenge, and propose   \textbf{the Pictorial Cortex (PictorialCortex)}, a framework that explicitly models the factors underlying cortical responses through a \emph{compositional latent formulation}.
Each fMRI observation is represented as a structured combination of four interpretable components:
(i) a \emph{stimulus-driven} factor encoding stable visual  content shared across subjects,
(ii) a \emph{dataset} factor capturing dataset- and protocol-specific variations,
(iii) a \emph{subject} factor reflecting individual cortical characteristics, and
(iv) a \emph{nuisance} factor modeling residual, trial-wise variability that does not carry stable or reusable stimulus semantics.
By organizing these components through conditioning and recomposition, PictorialCortex emphasizes stimulus-driven representations that can generalize across subjects, while accommodating subjective and experimental variability.
Importantly, this formulation operates on a \emph{universal cortical latent space} learned via a pretrained autoencoder, which maps heterogeneous fMRI activations into a compact, shared representation, providing a stable foundation for compositional modeling across subjects.

Formally, our PictorialCortex first pretrains a high-capacity autoencoder on cortical surface fMRI data from the large-scale UK Biobank~\cite{miller2016multimodal} dataset.
This pretraining embeds heterogeneous neural responses into a  shared and compact latent space that captures  cortical structure while abstracting away subject-specific variability.
This 
resulting \emph{universal cortical latent space} serves as a stable representational substrate across individuals, forming the foundation for downstream cross-subject modeling.

Building on this universal latent, we introduce a \emph{Latent Factorization--Composition Module} (LFCM) that factorizes each fMRI observation into four interpretable components: \emph{stimulus-driven visual content}, \emph{subject}, \emph{dataset}, and \emph{nuisance}. The Factorizer maps the universal latent to stimulus and nuisance codes conditioned on dataset and subject embeddings, while the Compositor recombines these components into surrogate latents. 
Notably, as  in Figure~\ref{fig:fmri_diff}(c), this factorization produces a stimulus-driven latent space that effectively aligns representations of the same stimulus across subjects while separating different stimuli within a single subject, capturing the objective visual content underlying the fMRI responses.
To encourage robust factorization, we employ two complementary training mechanisms: 
(i) \emph{Paired Factorization and Reconstruction} encourages paired fMRI observations of the same stimulus to share a stimulus-driven code while capturing residual variability in the nuisance component, and 
(ii) \emph{Re-Factorizing Consistency Regularization} exposes the Factorizer of LFCM to surrogate latents generated by the Compositor, encouraging that re-factorizing these latents recovers the original stimulus-driven and nuisance factors.

During inference for an unseen subject, we first extract an initial stimulus-driven code using a default subject embedding. To mitigate  distributional shifts,
the Compositor in LFCM synthesizes surrogate fMRI latents under multiple seen-subject conditions,
which are then re-factorized and aggregated to produce the final stimulus-driven code. 
This code subsequently guides diffusion-based image synthesis, enabling zero-shot cross-subject visual reconstruction.

Extensive experiments demonstrate that PictorialCortex substantially improves zero-shot cross-subject visual reconstruction, validating the effectiveness of compositional latent modeling for subject-agnostic neurodecoding. Furthermore, results show that our model benefits from multi-dataset training on UniCortex-fMRI, highlighting the importance of diverse subjects for robust cross-subject generalization.

\noindent\textbf{Contributions.}
(1) We construct UniCortex-fMRI, the unified cortical-surface fMRI dataset, integrating four visual-stimulus datasets with extensive subject and stimulus coverage, enabling systematic evaluation of zero-shot cross-subject generalization.
(2) We propose PictorialCortex, a compositional latent framework that disentangles fMRI signals into stimulus-driven, subject, dataset, and nuisance components, and employs a factorization–composition learning strategy with paired factorization and re-factorizing consistency regularization to encourage robust, invariant stimulus representations across subjects while modeling individual and experimental variability.
(3) We leverage UK Biobank pretraining to learn a universal cortical latent space, supporting multi-dataset, multi-subject learning without requiring subject-specific encoders.
(4) We achieve state-of-the-art zero-shot cross-subject fMRI-to-image reconstruction, demonstrating the effectiveness of compositional modeling, surrogate-latent aggregation, and dataset diversity for subject-agnostic neurodecoding.

\section{Related Work}
\label{sec:related}

\subsection{fMRI-to-image Reconstruction} 
This task focuses on reconstructing visual images from recorded fMRI signals. Most methods operate in a within-subject paradigm, training separate models for each individual to account for subject-specific neural response patterns.
Early fMRI decoding studies focused on coarse semantic prediction~\cite{cox2003functional}, image classification~\cite{haxby2001distributed, thirion2006inverse}, stimulus retrieval~\cite{kay2008identifying}, or reconstructing simple patterns such as handwritten digits~\cite{schoenmakers2013linear}. The advent of deep generative models, including GAN and VAE-based approaches, improved reconstruction fidelity for natural images~\cite{shen2019deep, mozafari2020reconstructing,ren2021reconstructing, gu2022neurogen}.
Recently, diffusion models~\cite{rombach2022high} have substantially advanced fMRI-to-image reconstruction~\cite{chen2023seeing, ozcelik2023natural, scotti2023reconstructing, zeng2023controllable, ferrante2023brain, fang2024alleviating}.
Chen et al.~\cite{chen2023seeing} leveraged masked brain modeling with Latent Diffusion Models to improve semantic alignment, though text-only conditioning limits spatial fidelity.
 Other approaches integrate CLIP features with VAE intermediates~\cite{ozcelik2023natural, scotti2023reconstructing, liu2025see, quan2024psychometry, wang2024unibrain} or incorporate low-level geometric cues like depth~\cite{ferrante2023brain} and silhouettes~\cite{zeng2023controllable} to enhance visual coherence.
Our conference version, NeuroPictor~\cite{neuropictor}, directly modulates diffusion model using multi-level fMRI conditioning, achieving precise control over semantic and spatial content. However, like most existing methods~\cite{neuropictor, scotti2024mindeye2, chen2022seeing, scotti2023reconstructing, ozcelik2023natural}, it is limited to within-subject evaluation. In contrast, this work addresses the zero-shot cross-subject fMRI-to-image reconstruction problem, where no test-subject-specific training data is available.

\subsection{Cross-subject fMRI Decoding} 
NeuroPictor~\cite{neuropictor} demonstrated that multi-subject pretraining followed by subject-specific refinement can improve within-subject decoding, highlighting the benefits of knowledge transfer across individuals. 
Similar observations have been reported in several NSD-based studies~\cite{scotti2024mindeye2, gong2025mindtuner, liu2025see, quan2024psychometry}. 
Beyond simple fine-tuning, more recent approaches~\cite{scotti2024mindeye2, wang2024mindbridge, gong2025mindtuner, han2024mindformer, xu2025cross} explore few-shot adaptation, where a small number of samples from a new subject are used to personalize a pretrained decoder.
Despite these advances, zero-shot cross-subject generalization remains largely unsolved. Kong et al.~\cite{ICLR2025_hcpcross} show on 177 HCP-Movie subjects that decoding performance for unseen individuals strongly depends on the number of training subjects, emphasizing subject diversity as a key limiting factor. However, existing datasets~\cite{allen2022massive,chang2019bold5000,nod2023,hcp2013wu} exhibit a fundamental imbalance: some provide dense stimulus coverage for only a few subjects, while others include many subjects but with extremely sparse stimulus exposure. This mismatch prevents systematic evaluation of zero-shot cross-subject fMRI-to-image reconstruction.
In this work, we introduce UniCortex-fMRI, a unified dataset specifically designed as a principled testbed for zero-shot cross-subject fMRI-to-image reconstruction. By consolidating multiple major datasets, including NSD~\cite{allen2022massive}, BOLD5000~\cite{chang2019bold5000}, NOD~\cite{nod2023}, and HCP-Movie~\cite{hcp2013wu}, UniCortex-fMRI jointly expands subject diversity and stimulus coverage, enabling controlled evaluation of cross-subject generalization.

\subsection{Pretrained fMRI Encoders} 
Learning robust fMRI encoders is crucial for  scalable and generalizable decoding.
Population-scale modeling efforts~\cite{caro2023brainlm, dong2024brain} capture high-level cohort features for tasks such as sex classification, age regression, or disease prediction, but these representations lack the spatial and semantic fidelity needed for fine-grained image reconstruction. 
Many methods also struggle with anatomical heterogeneity in voxel space, which makes representations difficult to transfer across individuals.
For example, masked brain modeling methods~\cite{chen2022seeing, chen2023cinematic} pretrained on BOLD5000 capture contextual structure but typically rely on subject-specific projections to accommodate varying voxel dimensionalities.
Similarly, recent multi-subject decoders~\cite{scotti2024mindeye2, wang2024mindbridge, gong2025mindtuner, han2024mindformer} rely on subject-specific linear projections to map heterogeneous voxel spaces into latent space.
In contrast, our work builds upon the fMRI autoencoding strategy introduced in NeuroPictor~\cite{neuropictor,qian2023fmri}, which maps  native-space fMRI signals onto a unified 2D fsLR cortical surface representation, enabling multi-individual modeling without subject-specific encoders.
PictorialCortex further extends this approach by large-scale pretraining on UK Biobank and subsequent factorization to the UniCortex-fMRI dataset. Importantly, in our framework, this unified encoder does not serve as the final representation, but as a stable foundation upon which explicit compositional modeling of stimulus-related representations under subject-, dataset-, and nuisance-related influences is performed to support robust zero-shot cross-subject decoding.

\subsection{fMRI-conditioned Diffusion Models}
Existing fMRI-conditioned diffusion methods mainly follow two paradigms. One line~\cite{scotti2024mindeye2,ozcelik2023natural, scotti2023reconstructing} maps fMRI signals into pretrained multimodal spaces, such as VAE latents, CLIP text and vision embeddings~\cite{radford2021learning}, and performs image-to-image diffusion~\cite{xu2023versatile} generation under these priors. While effective, their high-dimensional latent spaces introduce redundancy, increasing overfitting risk and limiting cross-subject generalization. 
Another line~\cite{neuropictor,chen2023seeing}, relies on end-to-end implicit learning without explicit feature alignment, which lacks unified supervision and can be unstable for zero-shot cross-subject reconstruction. In contrast, we adopt IP-Adapter~\cite{ipadapter} as it operates on a compact latent space and enables direct image reconstruction from a single feature, making it particularly suited for robust zero-shot cross-subject fMRI-to-image reconstruction.

\section{Task and Dataset}

\begin{figure}
\centering
\includegraphics[width=\linewidth]{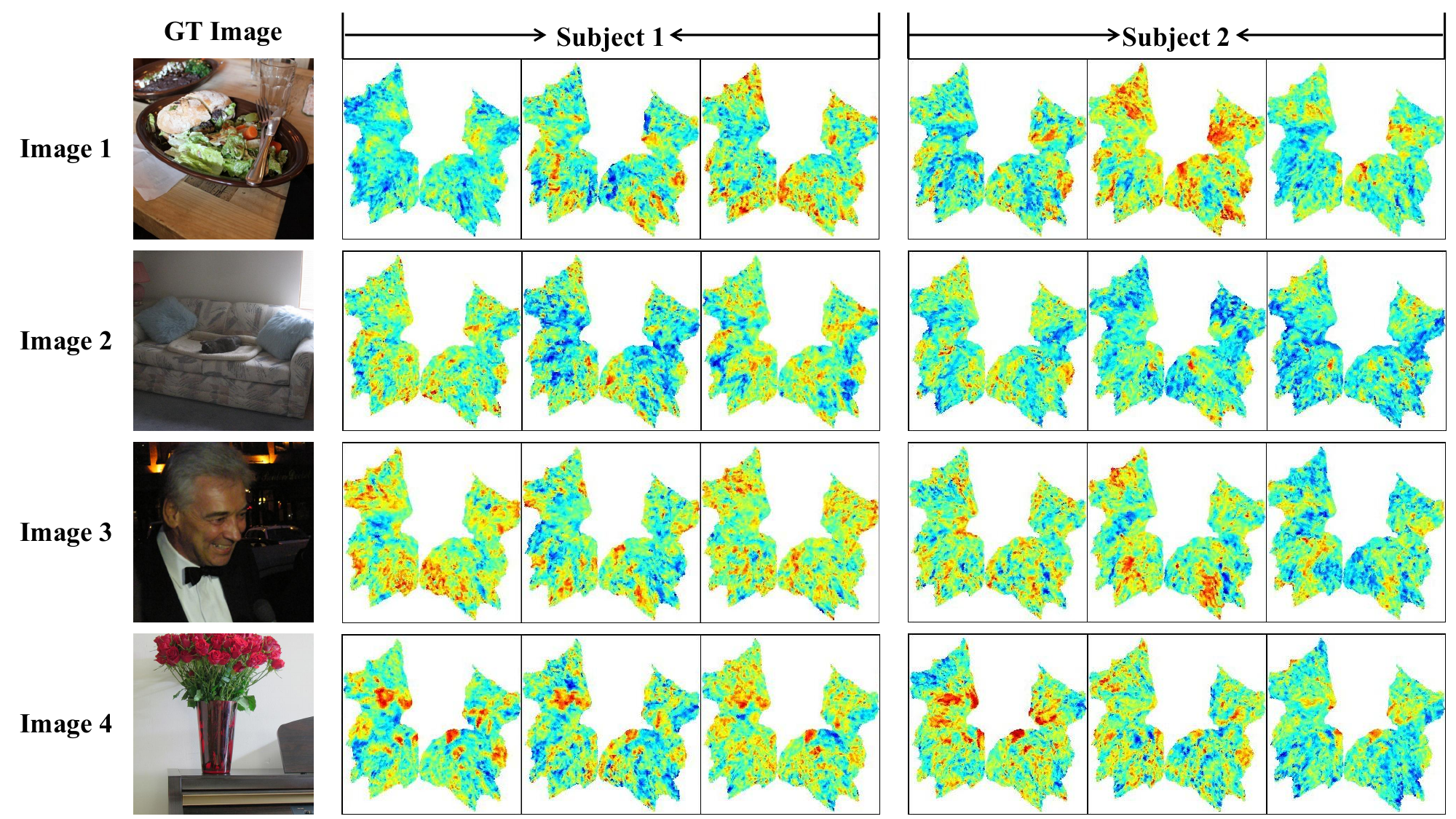}
\caption{
Visualizations of fMRI responses from two subjects viewing the same visual stimuli.
In each row, the first column shows the ground-truth image.
Columns 2--4 display three fMRI responses from Subject~1 under repeated presentations of the same image
, while columns 5--7 show the corresponding three fMRI responses from Subject~2. Full size is in Appendix.
}
\label{fig:supp_fmri}
\end{figure}

\subsection{Task definition}

We study the problem of visual neurodecoding, which aims to reconstruct a perceived visual stimulus from human fMRI measurements.
Let $\mathbf{S}$ denote an fMRI observation recorded from a subject during the presentation of a visual stimulus, and let $\mathbf{I}$ denote the ground-truth image.
Each fMRI sample is additionally associated with a subject identity $\mathrm{sub}$ and a dataset or acquisition condition $\mathrm{dataset}$, reflecting differences in experimental protocols, scanners, and stimulus paradigms.

The goal of visual neurodecoding is to learn a mapping
\begin{equation}
f: \mathbf{S} \rightarrow \mathbf{I},
\end{equation}
such that the reconstructed image faithfully reflects the visual content that elicited the observed neural response.

\noindent\textbf{Within-subject decoding.}
Most existing fMRI-to-image reconstruction methods operate in a within-subject setting, where both training and testing samples are drawn from the same individual under fixed experimental conditions.
In this regime, subject- and trial-specific variability is not explicitly modeled;
models implicitly assume such variability is negligible or constant. While this simplification facilitates learning, it results in models that are tightly coupled to individual subjects and experimental setups, limiting generalization to new subjects.

\noindent\textbf{Zero-shot cross-subject decoding.}
Thus, we tackle with \emph{zero-shot cross-subject visual decoding}.
Formally, during training the model observes fMRI samples from a set of subjects $\mathcal{U}_{\mathrm{train}}$, while at test time it is evaluated on samples from a disjoint set of unseen subjects $\mathcal{U}_{\mathrm{test}}$, where $\mathcal{U}_{\mathrm{train}} \cap \mathcal{U}_{\mathrm{test}} = \varnothing$.
The objective is to reconstruct the visual stimulus experienced by an unseen subject only from their fMRI measurements, without any subject-specific fine-tuning.

This setting highlights the non-injective nature of the stimulus–fMRI mapping:
identical visual stimuli  can evoke  highly variable  neural responses across subjects and trials due to anatomical, cognitive, and experimental differences.
As in Figure~\ref{fig:supp_fmri}, even for the same image, fMRI responses can differ markedly between subjects and across repeated trials within a single subject.
Effective zero-shot decoding therefore requires disentangling stimulus-driven visual content from subject-, dataset-, and nuisance-related factors embedded in fMRI signals.

\subsection{UniCortex-fMRI Dataset}

\begin{figure}[t]
\centering
\includegraphics[width=1.0\linewidth]{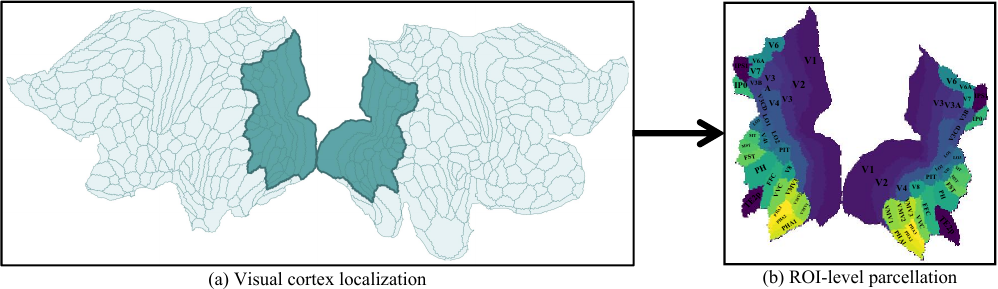}
\caption{
(a) Visualization of the visual cortex (VC) highlighted within the whole cortical surface.
(b) ROI-level parcellation within the visual cortex, where colors indicate different visual areas.}
\label{fig:vc_roi}
\end{figure}

Reliable evaluation of zero-shot cross-subject neurodecoding requires both
broad subject diversity and sufficient visual stimulus variability.
However, no prior resource jointly offers large-scale subject diversity and rich visual stimulus coverage suitable for visual decoding at scale. Existing fMRI datasets typically emphasize only on one of these dimensions:
datasets with rich stimulus coverage often include only a small number of subjects,
while datasets with many participants provide limited and highly constrained stimulus exposure.
This structural imbalance makes it difficult to disentangle stimulus-driven representations from dataset-,
subject-, and nuisance-related variability, and severely limits cross-subject generalization.

To address this limitation, we construct \textbf{UniCortex-fMRI},
\textit{a large dataset for visual-stimulus fMRI decoding with standardized preprocessing and unified data formats}, making the dataset directly and publicly usable by the research community.
UniCortex-fMRI consolidates multiple large-scale datasets into a unified representational and experimental framework, enabling joint modeling across heterogeneous acquisition protocols and systematic evaluation of zero-shot cross-subject decoding at scale.
Specifically, UniCortex-fMRI integrates four widely used datasets:
the Natural Scenes Dataset (NSD),
the Brain, Object, Landscape Dataset (BOLD5000),
the Natural Object Dataset (NOD),
and the Human Connectome Project movie-watching dataset (HCP-Movie).
These datasets cover complementary regimes of subject diversity and stimulus complexity.
In total, the dataset comprises 219 subjects and 816,660 trial-wise pairs of fMRI recordings and visual stimuli.

Across all datasets, raw fMRI recordings are reprocessed using a unified pipeline.
All data are projected onto the fsLR-32k cortical surface, normalized by per-session $z$-scoring, and restricted to the visual cortex. We select both early and higher-level visual cortex regions of interest (RoIs) from the HCP-MMP atlas in fsLR32K space, including
``V1, V2, V3, V3A, V3B, V3CD, V4, LO1, LO2, LO3, PIT, V4t, V6, V6A, V7, V8, PH, FFC, IP0, MT, MST, FST, VVC, VMV1, VMV2, VMV3, PHA1, PHA2, PHA3, TE2p, IPS1''
(Figure~\ref{fig:vc_roi}). 
The resulting signals are converted into standardized $256 \times 256$ cortical activation maps.
This harmonized representation enables consistent definition of training and testing splits, including controlled seen/unseen subject partitions, and supports joint learning across datasets without subject-specific encoders.

Below, we describe the dataset-specific characteristics and train/test splits used throughout this work.

%-------------------------------------------------------------------------
% \subsubsection{Natural Scenes Dataset}
\noindent\textbf{Natural Scenes Dataset.}
NSD presents 73,000 natural images sourced from the MS-COCO dataset~\cite{lin2014microsoft}. 
In the conference version of NeuroPictor~\cite{neuropictor}, only the earlier incomplete NSD release was used. 
Here, we additionally incorporate the newly released three sessions for each subject. 
NSD includes eight subjects: subjects 1, 2, 5, 7 each viewed 10,000 images three times (30,000 trials per subject); 
subjects 3 and 6 viewed 8,000 images three times (24,000 trials per subject); 
subjects 4 and 8 viewed 7,500 images three times (22,500 trials per subject).  
Following NSD's standard protocol, the 1,000 images viewed by all subjects are used as the test set. 
We further designate subject 1 as an unseen subject excluded from all training for evaluating zero-shot cross-subject decoding.

%-------------------------------------------------------------------------
% \subsubsection{HCP-Movie}
\noindent\textbf{HCP-Movie.}
The HCP-Movie dataset includes 177 participants watching four movie clips (about 15 minutes each). 
In HCP-Movie, fMRI is natively sampled at 1~Hz, allowing each volume to be aligned with the video frame shown in the corresponding second. A 4-second shift is applied to compensate for the hemodynamic delay.
Following the data split protocol in \cite{ICLR2025_hcpcross}, we remove blank frames and obtain 3,027 frames for training and 100 frames for testing for each subject.
Subjects 1--10 are held out as unseen subjects.

%-------------------------------------------------------------------------
% \subsubsection{BOLD5000}
\noindent\textbf{BOLD5000.}
BOLD5000 includes four subjects who collectively viewed almost 5,000 images: 
1,000 scene images, 2,000 COCO images, and 1,916 ImageNet images. 
We found that a subset of COCO images overlaps with the NSD test images. 
To avoid inadvertent cross-dataset leakage, all overlapping COCO images are reassigned to the test split.
This results in 3,923 training images and 1,331 test images, differing from the official split. 
Subject~1 is designated as the unseen subject.

%-------------------------------------------------------------------------
\noindent\textbf{Natural Object Dataset.}
NOD includes 30 participants and a total of 57,620 naturalistic images.
Among the 30 participants, nine subjects (subjects 1–9) viewed a richer set of stimuli:
they each completed four non-overlapping ImageNet sessions (4,000 unique images per subject) plus a COCO session containing 120 identical images shared across all nine subjects, each repeated ten times.
The remaining 21 subjects (subjects 10–30) completed only a single ImageNet session with 1,000 subject-specific images.
For experimental consistency, we construct subject-wise train/test splits as follows:
for subjects 1–9, we use the 10 shared COCO images and 200 ImageNet images per subject as the test set;
for subjects 10–30, we select 50 images per subject as the test set.
All remaining images are used for training.
We designate subject 1 as an unseen subject for evaluating cross-subject generalization.

%-------------------------------------------------------------------------
\section{Method}

\begin{figure*}[!t]
\centering
\includegraphics[width=\linewidth]{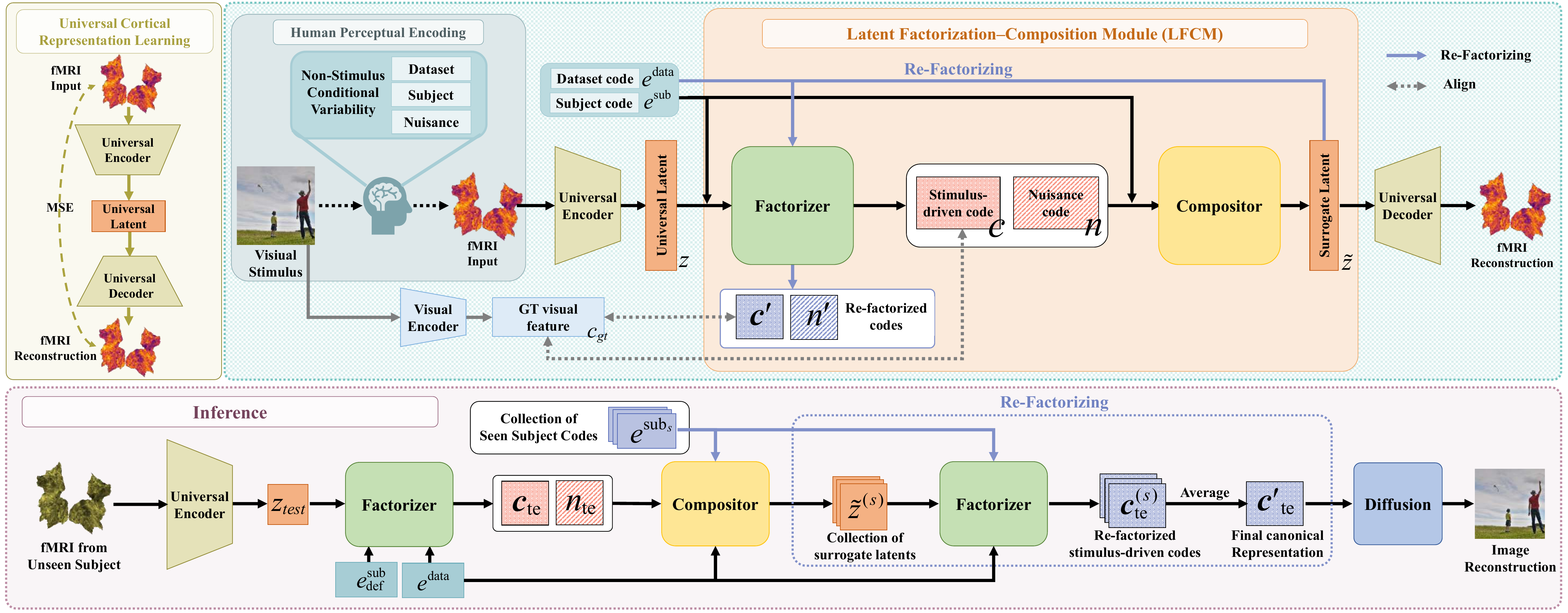}
\caption{Overview of the PictorialCortex framework. 
Given an fMRI signal arising from a visual stimulus, the observed cortical response is jointly influenced by stimulus-driven visual content as well as non-stimulus variability, including subject identity, dataset context, and trial-wise nuisance. 
(i) \textbf{Top-left:} Universal cortical representation learning maps fMRI inputs to a shared representational space, enabling reconstruction across subjects.
(ii) \textbf{Top-right:} We first encode the fMRI input into a universal fMRI latent $\mathbf{z}$ using the universal encoder. 
Then, we perform \emph{compositional latent modeling} via the proposed Latent Factorization--Composition Module (LFCM).
The Factorizer decomposes each universal latent into a stimulus-driven code $\mathbf{c}$ and a nuisance code $\mathbf{n}$, conditioned on subject and dataset codes.
These components are recombined by the Compositor to synthesize surrogate fMRI latents $\tilde{\mathbf{z}}$.
A subsequent re-factorization step is trained to enforce consistency when re-encoding surrogate fMRI latents.
Both the original stimulus-driven code $\mathbf{c}$ and the re-factorized stimulus-driven code $\mathbf{c}'$ are aligned with the ground-truth visual target $\mathbf{c}_{\mathrm{gt}}$,
while the nuisance codes $\mathbf{n}$ and $\mathbf{n}^{\prime}$ are aligned to encourage consistency under re-factorization.
(iii) \textbf{Bottom:} During inference, fMRI signals from unseen subjects are factorized, composited and re-factorized to produce a refined stimulus-driven code (\( \mathbf{c}^{\prime}_{\mathrm{te}} \)) using seen subject conditions, which conditions the diffusion model to generate the reconstructed image (\( \hat{\mathbf{I}} \)).
}
\label{fig:pipeline}
\end{figure*}

We build \textbf{PictorialCortex} on the UniCortex-fMRI dataset. 
As  in Figure~\ref{fig:pipeline}, it employs a \emph{compositional latent formulation} that factorizes each fMRI observation into stimulus-driven visual, subject, dataset, and nuisance components.
This design isolates invariant visual information while capturing residual, trial-wise variability in the nuisance factor, enabling robust zero-shot cross-subject decoding.

\noindent \textbf{Overview}. 
As in Fig.~\ref{fig:pipeline}, our PictorialCortex consists of three stages.
(1) \textit{Stage I: Universal Cortical Representation Learning.}
We pretrain a high-capacity autoencoder on large-scale cortical fMRI data to obtain a universal latent space, providing a stable, shared domain for modeling heterogeneous observations across subjects (Sec.~\ref{sec:universal_latent}).
(2) \textit{Stage II: Compositional Latent Modeling.}
We factorize each fMRI observation into stimulus-driven, subject, dataset, and nuisance components using the proposed Latent Factorization--Composition Module (LFCM).
This stage isolates invariant visual content while explicitly modeling individual-specific characteristics, dataset-dependent context, and residual trial-wise variability.
Robustness is enforced via Paired Factorization and Reconstruction (PFR) and Re-Factorizing Consistency Regularization (ReFCR) (Sec.~\ref{sec:lfcm}--\ref{sec:refcr}).
(3) \textit{Stage III: Inference.}
At inference, the stimulus-driven code of an unseen subject is refined via surrogate re-factorization across multiple seen-subject conditions, and subsequently used to condition a diffusion model for zero-shot image reconstruction (Sec.~\ref{sec:reconstruction}).

%-------------------------------------------------------------------------
\subsection{Universal Cortical Representation Learning}
\label{sec:universal_latent}

To enable the proposed compositional latent formulation and support reliable cross-subject modeling, we first learn a \emph{universal cortical latent representation} shared across individuals.
This stage aims to establish a common representational domain for heterogeneous fMRI observations, providing a stable and anatomically consistent foundation for subsequent factorization.

 \noindent\textbf{Architecture.} We train a high-capacity autoencoder on cortical surface fMRI data to map raw surface-level activations into a compact universal latent space that captures shared cortical structure while preserving trial-specific variability, providing a consistent anatomical reference across individuals and experimental settings. The autoencoder takes as input a cortical surface map $\mathbf{S} \in \mathbb{R}^{256 \times 256}$; anatomically empty regions are removed prior to patch embedding to eliminate redundant spatial tokens and focus capacity on informative cortical areas. The remaining patches, together with learnable CLS tokens, are processed by a deep transformer encoder ($\mathcal{E}_{\mathcal{A}}(\cdot)$) to produce the latent representation,
\begin{equation}
\mathbf{z} = \mathcal{E}_{\mathcal{A}}(\mathbf{S}),
\end{equation}
where $\mathbf{z} \in \mathbb{R}^{L_r \times d_r}$, $L_r$ denotes the number of CLS tokens, and $d_r$ their feature dimension.
This latent representation serves as a \emph{universal cortical embedding} prior to any explicit compositional modeling.

A symmetric transformer decoder  reconstructs the cortical surface map by injecting the CLS tokens and replacing masked spatial tokens with learnable guide tokens. This design forces the encoder to compress the entire cortical activation pattern into a compact set of latent tokens, encouraging the model to learn expressive representations.
The decoder ($\mathcal{D}_{\mathcal{A}}(\cdot)$) outputs patch embeddings, which are mapped back to pixel space via an unpatchify operation. So  we have the reconstructed cortical surface
\begin{equation}
\tilde{\mathbf{S}} = \mathcal{D}_{\mathcal{A}}(\mathbf{z}).
\end{equation}

The autoencoder is trained using a reconstruction objective that minimizes the mean squared error between the original and reconstructed cortical surface maps:
\begin{equation}
\mathcal{L}_{\mathcal{A}} = \left\lVert \mathbf{S} - \tilde{\mathbf{S}} \right\rVert_2^2,
\end{equation}
where $\mathbf{S}$ and $\tilde{\mathbf{S}}$ denote the ground-truth and reconstructed fMRI surface maps, respectively. We pretrain this autoencoder on large-scale resting-state fMRI data from more than 40,000 subjects in the UK Biobank dataset~\cite{miller2016multimodal}.
This large-scale pretraining yields a stable and expressive cortical latent space that generalizes across individuals and datasets.

%-------------------------------------------------------------------------
\subsection{Latent Factorization--Composition Module (LFCM)}
\label{sec:lfcm}

The goal of compositional latent modeling is to derive a structured latent formulation that separates stimulus-driven information from heterogeneous sources of variability inherent in multi-dataset, multi-subject fMRI data. 

\subsubsection{Method Overview}
\label{sec:lfcm_overview}

LFCM operates on the universal cortical latent representation $\mathbf{z}$ learned in Sec.~\ref{sec:universal_latent}.
We introduce learnable embeddings $\mathbf{e}^{\mathrm{sub}}$ and $\mathbf{e}^{\mathrm{data}}$ to represent subject identity and dataset context, respectively, which serve as conditioning signals during latent factorization and recomposition.
Conditioned on subject and dataset context, each fMRI observation is factorized into
(i) a stimulus-driven latent $\mathbf{c}$ encoding stable visual content shared across subjects, and
(ii) a nuisance latent $\mathbf{n}$ capturing residual variability that is not transferable across subjects.

Specifically, LFCM consists of two complementary components:
a \emph{Factorizer} $\mathcal{F}$ that decomposes a universal latent into
structured components, and a \emph{Compositor} $\mathcal{C}$ that
recombines these components back into the universal latent space.

\noindent \textbf{Factorizer.}
Given a universal fMRI latent $\mathbf{z}$, the Factorizer ($\mathcal{F}$) performs
subject- and dataset-conditioned latent factorization to parse the representation
\begin{equation}
(\mathbf{c}, \mathbf{n}) =
\mathcal{F}(\mathbf{z} \mid \mathbf{e}^{\mathrm{sub}}, \mathbf{e}^{\mathrm{data}}), \label{eq_c_n}
\end{equation}
into a stimulus-driven component $\mathbf{c}$ and a residual nuisance component $\mathbf{n}$. And $\mathbf{e}^{\mathrm{sub}}$ and $\mathbf{e}^{\mathrm{data}}$ are learnable embeddings
encoding subject identity and dataset context, respectively.
The output $\mathbf{c}$ is encouraged to capture stimulus-consistent
visual structure, while $\mathbf{n}$ absorbs residual variability that is
not reusable across subjects.

Concretely, the universal latent $\mathbf{z}$ is first projected into an
intermediate representation,
which is then modulated by additive subject and dataset embeddings, 
in analogy to positional encodings in transformer,
\begin{equation}
\mathbf{h} =  \mathrm{Linear}(\mathbf{z}) + \mathbf{\mathbf{e}^{\mathrm{sub}}} + \mathbf{\mathbf{e}^{\mathrm{data}}}.
\end{equation}
During training, a small fraction (5\%) of subject embeddings are replaced
with a default subject embedding $\mathbf{e}^{\mathrm{sub}}_{\mathrm{def}}$ (in Eq.~(\ref{eq_default})), serving as the `placeholder' for a neutral subject 
without individual-specific information in the inference stage.
This enables the Factorizer to encode stimulus-driven content under unknown or unseen subject conditions.
The conditioned representation $\mathbf{h}$ is then processed by a shared
transformer stack with learnable queries to extract the stimulus-driven
latent $\mathbf{c}$ and the nuisance latent $\mathbf{n}$.

\noindent \textbf{Compositor.}
The Compositor ($\mathcal{C}$) reconstructs a surrogate fMRI universal latent representation 
\begin{equation}
\tilde{\mathbf{z}} =
\mathcal{C}(\mathbf{c}, \mathbf{n} \mid \mathbf{e}^{\mathrm{sub}}, \mathbf{e}^{\mathrm{data}}).
\end{equation}
Here, the stimulus-driven and nuisance latents ($\mathbf{c}$ and $\mathbf{n}$) in Eq.~(\ref{eq_c_n}) are  concatenated and added by subject and dataset embeddings ($\mathbf{e}^{\mathrm{sub}}$, $\mathbf{e}^{\mathrm{data}}$). It is further processed by a transformer with learnable queries that attend to these conditioned latents to produce the reconstructed surrogate fMRI universal latent. The network details are in Supplementary Material.

This factorization--composition design enables controlled recombination of stimulus, subject, dataset and nuisance information, forming the basis for both training and inference.

\subsection{Training Mechanisms}

To ensure a stable and interpretable compositional structure, we employ two complementary strategies:

\noindent (i) \emph{Paired Factorization and Reconstruction (PFR)} ensures that the Factorizer and Compositor can faithfully decompose and recompose real fMRI latents under matched subject and dataset conditions:
\begin{equation}
\mathbf{z}
\xrightarrow{\mathcal{F}(\cdot \mid \mathbf{e}^{\mathrm{sub}}, \mathbf{e}^{\mathrm{data}})}
(\mathbf{c}, \mathbf{n})
\xrightarrow{\mathcal{C}(\cdot \mid \mathbf{e}^{\mathrm{sub}}, \mathbf{e}^{\mathrm{data}})}
\tilde{\mathbf{z}}.
\end{equation}

\noindent  (ii) \emph{Re-Factorizing Consistency Regularization (ReFCR)} further constrains the Factorizer by re-encoding recomposed surrogate latents to recover consistent stimulus-driven and nuisance components:
\begin{equation}
\tilde{\mathbf{z}}
\xrightarrow{\mathcal{F}(\cdot \mid \mathbf{e}^{\mathrm{sub}}, \mathbf{e}^{\mathrm{data}})}
(\mathbf{c}', \mathbf{n}').
\end{equation}
This encourages the Factorizer to be robust and preserves stimulus-driven semantics even beyond the observed fMRI manifold. We give the details of PFR and ReFCR next.

\begin{figure*}[!t]
\centering
\includegraphics[width=\linewidth]{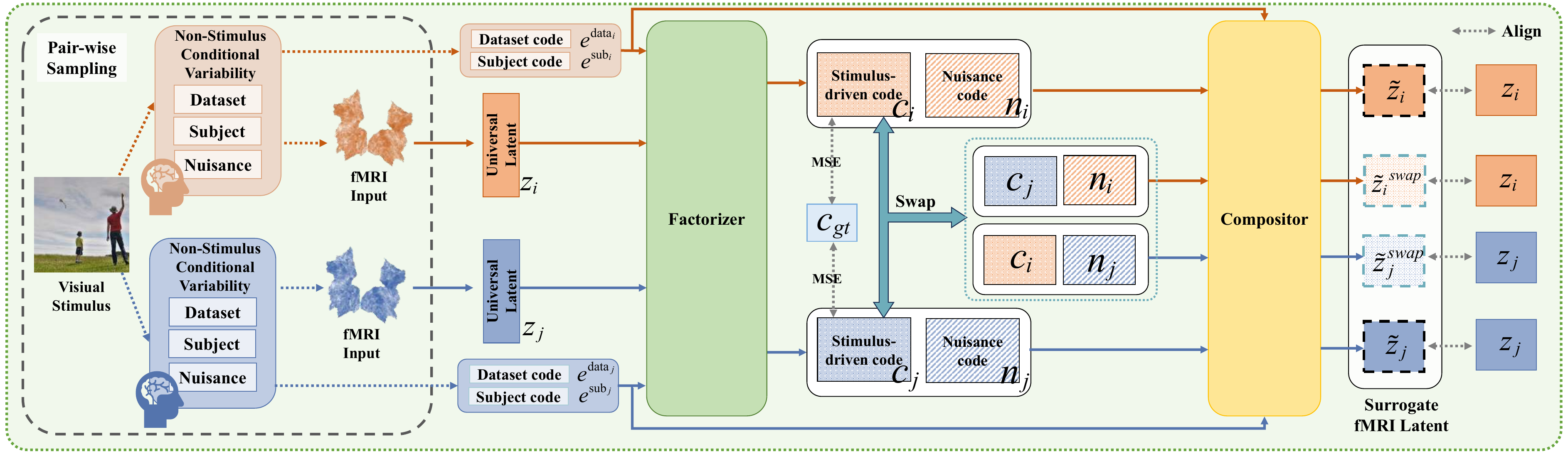}
\caption{Overview of Paired Factorization and Reconstruction (PFR).
Paired fMRI observations elicited by the same visual stimulus are encoded into a universal latent space and decomposed into stimulus-driven and nuisance components under subject and dataset conditioning.
The stimulus-driven components from the paired samples are aligned to a shared visual target.
To further strengthen disentanglement, a pairwise swapping operation exchanges the stimulus-driven codes between paired observations while preserving the other factors, and consistency is enforced by reconstructing the original fMRI signals both with and without swapping.
Together, PFR constrains the stimulus-driven latent to be invariant across paired observations, while allowing non-stimulus factors to account for other variability.
}
\label{fig:prf}
\end{figure*}

\subsubsection{Paired Factorization and Reconstruction}
\label{sec:prf}

Paired Factorization and Reconstruction (PFR) aims to learn a stable \emph{compositional latent formulation} by leveraging multiple fMRI observations elicited by the same visual stimulus.
The core intuition is that, although fMRI signals vary across trials or subjects, the underlying stimulus-driven information should remain invariant.
PFR explicitly enforces this invariance by encouraging paired observations to share a common stimulus-driven representation, while attributing residual variability to nuisance factors.

\noindent\textbf{Implementations.}
Consider a paired set of fMRI surface maps $(\mathbf{S}_i, \mathbf{S}_j)$ corresponding to the same visual stimulus.
Paired samples may come from repeated trials of the same subject or from different subjects, while typically sharing the same dataset context.
Each surface map is first mapped into the universal cortical latent space (Sec.~\ref{sec:universal_latent}), producing universal latents $\mathbf{z}_i$ and $\mathbf{z}_j$.

The Latent Factorization--Composition Module then applies the \emph{Factorizer} to decompose each universal latent into a stimulus-driven component and a nuisance component:
\begin{align}
(\mathbf{c}_i, \mathbf{n}_i) &= \mathcal{F}(\mathbf{z}_i \mid \mathbf{e}^{\mathrm{sub}_i}, \mathbf{e}^{\mathrm{data}_i}), \\
(\mathbf{c}_j, \mathbf{n}_j) &= \mathcal{F}(\mathbf{z}_j \mid \mathbf{e}^{\mathrm{sub}_j}, \mathbf{e}^{\mathrm{data}_j}).
\end{align}

To enforce consistency of stimulus-driven representations across paired
observations, we align $\mathbf{c}_i$ and $\mathbf{c}_j$ to a shared visual
target representation $\mathbf{c}^{\mathrm{gt}}$ using an $\ell_2$ objective:
\begin{equation}
\mathcal{L}_{\mathrm{align}}
=
\sum_{k \in \{i,j\}}
\left\lVert \mathbf{c}_k - \mathbf{c}^{\mathrm{gt}} \right\rVert_2^2 .
\end{equation}
 Here, the ground-truth visual target $\mathbf{c}^{\mathrm{gt}}$ is obtained by extracting image feature from the original image using IP-Adapter and its internal resampler. This alignment encourages the stimulus-driven components to encode identical visual semantics, independent of subject- or trial-specific variability.

Given the decomposed components, the Compositor recombines them to
reconstruct each universal latent under its corresponding subject and dataset embeddings:
\begin{equation}
\tilde{\mathbf{z}}_k =
\mathcal{C}(\mathbf{c}_k, \mathbf{n}_k
\mid \mathbf{e}^{\mathrm{sub}_k}, \mathbf{e}^{\mathrm{data}_k}),
\qquad k \in \{i,j\}.
\end{equation}

Reconstruction fidelity is enforced jointly in the universal latent space
and the original fMRI surface space:
\begin{equation}
\mathcal{L}_{\mathrm{rec}_k}
=
\left\lVert \mathbf{z}_k - \tilde{\mathbf{z}}_k \right\rVert_2^2
+
\left\lVert
\mathbf{S}_k -
\mathcal{D}_{\mathcal{A}}(\tilde{\mathbf{z}}_k)
\right\rVert_2^2,
\qquad k \in \{i,j\}.
\end{equation}

Importantly, we introduce a \emph{pairwise swapping} operation that enforces stimulus invariance with respect to subject and trial factors, mitigating spurious correlations arising from subject- and trial-specific variability.
Specifically, for each paired sample $(\mathbf{z}_i, \mathbf{z}_j)$, we recombine the stimulus-driven component
from one observation with the nuisance component and conditioning factors from
the other.
Without loss of generality, we construct:
\begin{equation}
\tilde{\mathbf{z}}_{i}^{\mathrm{swap}} =
\mathcal{C}(\mathbf{c}_j, \mathbf{n}_i
\mid \mathbf{e}^{\mathrm{sub}_i}, \mathbf{e}^{\mathrm{data}_i}),
\end{equation}
and define $\tilde{\mathbf{z}}_{j}^{\mathrm{swap}}$ analogously by exchanging
the roles of $i$ and $j$. Thus, the  reconstruction consistency under pairwise swapping is enforced by:
\begin{equation}
\mathcal{L}_{\mathrm{rec_{swap}}}
=
\sum_{k \in \{i,j\}}
\Big(
\left\lVert \mathbf{z}_k - \tilde{\mathbf{z}}_{k}^{\mathrm{swap}} \right\rVert_2^2
+
\left\lVert
\mathbf{S}_k -
\mathcal{D}_{\mathcal{A}}(\tilde{\mathbf{z}}_{k}^{\mathrm{swap}})
\right\rVert_2^2
\Big).
\end{equation}
This operation enforces disentanglement by ensuring that $\mathbf{c}$ encodes invariant visual semantics across paired observations, while the other codes encode the dataset-, subject- and residual trial-specific variability.
Thus, the overall reconstruction objective is:
\begin{equation}
\mathcal{L}_{\mathrm{rec}}
=
\mathcal{L}_{\mathrm{rec}_i}
+
\mathcal{L}_{\mathrm{rec}_j}
+
\mathcal{L}_{\mathrm{rec_{swap}}}.
\end{equation}

\subsubsection{Re-Factorizing Consistency Regularization}
\label{sec:refcr}

Paired Factorization and Reconstruction (PFR) enforces disentanglement only on  \emph{real} fMRI latents \emph{on-manifold} observed during training, providing limited guarantees under distribution shift.
In contrast, zero-shot cross-subject decoding is inherently \emph{off-manifold}, as neural responses from unseen subjects may deviate systematically from the empirical training distribution.
We therefore introduce \emph{Re-Factorizing Consistency Regularization} (ReFCR) to promote stable factorization beyond the observed manifold.
ReFCR leverages surrogate fMRI latents synthesized by the Latent Factorization–Composition Module (LFCM) Compositor, which recombines stimulus and nuisance factors to generate structurally valid but unobserved neural configurations.
Enforcing consistent re-factorization on these surrogate latents effectively expands the Factorizer’s training domain and improves off-manifold generalization.

\noindent \textbf{Implementations}.
Given a surrogate latent $\tilde{\mathbf{z}}^{sg}_i$, either reconstructed from factorized components ($\tilde{\mathbf{z}}_{i}$) or generated via pairwise swapping operation ($\tilde{\mathbf{z}}_{i}^{\mathrm{swap}}$), we re-encode it using the Factorizer conditioned on the corresponding subject and dataset embeddings:
\begin{equation}
(\mathbf{c}_i^{\prime}, \mathbf{n}_i^{\prime}) =
\mathcal{F}(\tilde{\mathbf{z}}^{sg}_i \mid \mathbf{e}^{\mathrm{sub}^{sg}_i}, \mathbf{e}^{\mathrm{data}^{sg}_i}).
\end{equation}
Let $\mathbf{n}^{sg}_i$ denote the nuisance component used to construct
$\tilde{\mathbf{z}}^{sg}_i$.
ReFCR enforces that re-factorizing surrogate latents
recovers a stimulus-driven representation aligned with the ground-truth visual target $\mathbf{c}^{\mathrm{gt}}_i$,
while preserving the original nuisance structure $\mathbf{n}^{sg}_i$.
The resulting consistency objective is
\begin{equation}
\mathcal{L}_{\mathrm{ReFCR}}
=
\sum_{i}
\left(
\left\lVert \mathbf{c}_i^{\prime} - \mathbf{c}^{\mathrm{gt}}_i \right\rVert_2^2
+
\left\lVert \mathbf{n}_i^{\prime} - \mathbf{n}^{sg}_i \right\rVert_2^2
\right).
\end{equation}

By enforcing compositional consistency on surrogate latents,
ReFCR trains the Factorizer to preserve stimulus semantics
and nuisance structure beyond the observed fMRI manifold,
thereby complementing the disentanglement learned by PFR and strengthening cross-subject generalization.

\noindent \textbf{Final Objectives}.
All components of the proposed framework are learned jointly by optimizing a sum of objectives:
\begin{equation}
\mathcal{L}
=
\mathcal{L}_{\mathrm{rec}}
+ \mathcal{L}_{\mathrm{align}}
+ \mathcal{L}_{\mathrm{ReFCR}}
\end{equation}

%-------------------------------------------------------------------------
\subsection{Inference}
\label{sec:reconstruction}

Given an unseen subject, our goal is to reconstruct the perceived visual stimulus
without any subject-specific fine-tuning. 
Let $\mathbf{z}_{\text{te}}$ denote the universal fMRI latent of a test sample.
We first perform a preliminary factorization using the LFCM Factorizer:
\begin{equation}
(\mathbf{c}_{\text{te}}, \mathbf{n}_{\text{te}}) =
\mathcal{F}(\mathbf{z}_{\text{te}} \mid \mathbf{e}^{\mathrm{sub}}_{\mathrm{def}}, \mathbf{e}^{\mathrm{data}}), \label{eq_default}
\end{equation}
where $\mathbf{e}^{\mathrm{sub}}_{\mathrm{def}}$ is a shared default subject embedding (Sec.~\ref{sec:lfcm_overview}) used for all unseen individuals. 

Although $\mathbf{c}_{\text{te}}$ provides an initial estimate of the stimulus-driven content,
directly decoding it can be suboptimal due to the distributional shifts between seen and unseen subjects. Therefore, to obtain a robust, subject-invariant representation, we use the Compositor as a latent-space fMRI generator, re-expressing the same stimulus under multiple subject-specific configurations.
Specifically, we combine the factorized test latents $(\mathbf{c}_{\text{te}}, \mathbf{n}_{\text{te}})$ with a collection of subject embeddings drawn from the corresponding training dataset, denoted by 
$\{\mathbf{e}^{\mathrm{sub}}_s\}_{s \in \mathcal{S}}$, where $\mathcal{S}$ indexes all subjects in the corresponding training set, as well as a default subject embedding.
This combination is used to synthesize a collection of surrogate fMRI latents 
$\{\tilde{\mathbf{z}}^{(s)}\}_{s \in \mathcal{S}}$. 
Each surrogate latent is then re-factorized by the Factorizer, producing a subject-conditioned stimulus-driven code $\mathbf{c}_{\mathrm{te}}^{(s)}$.

To align these codes with the training manifold, each $\mathbf{c}_{\text{te}}^{(s)}$ is normalized and rescaled using the mean and standard deviation of the training visual features under the corresponding acquisition condition, yielding a stable canonical latent for diffusion-based image synthesis.

The final stimulus representation is obtained by averaging over these re-factorized stimulus-driven codes across all subject-conditioned surrogates:
\begin{equation}
\mathbf{c}^{\prime}_{\text{te}}
=
\frac{1}{|\mathcal{S}|}
\sum_{s \in \mathcal{S}}
\mathbf{c}_{\text{te}}^{(s)}.
\end{equation}
Aggregating across subject-conditioned latents produces a canonical representation that mitigates subject bias and stabilizes predictions.

Finally, the canonical representation $\mathbf{c}^{\prime}_{\text{te}}$ is fed to the IP-Adapter diffusion model to generate a high-fidelity reconstruction of the perceived visual content.
\begin{equation}
\hat{\mathbf{I}} = \mathrm{Diffusion}\big(\mathbf{c}^{\prime}_{\text{te}}\big).
\end{equation}
The resulting image $\hat{\mathbf{I}}$ serves as the reconstructed visual stimulus corresponding to the input fMRI signal.

%-------------------------------------------------------------------------
\section{Experiments}

\begin{figure*}[!t]
\centering
\includegraphics[width=\linewidth]{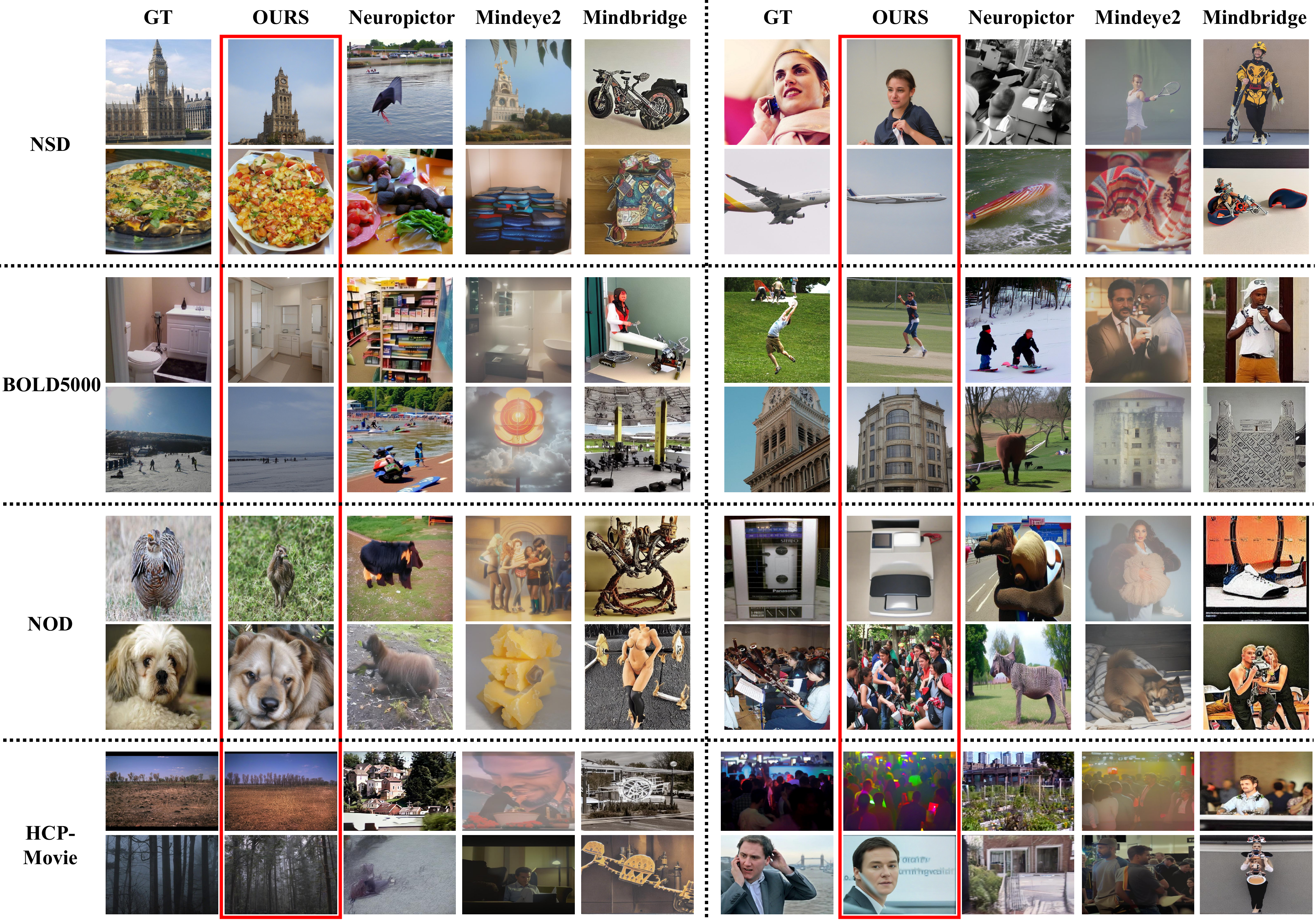}
\caption{Qualitative comparison of our method and representative competitors. 
Rows 1--2: NSD; rows 3--4: BOLD5000; rows 5--6: NOD; rows 7--8: HCP-movie. 
For each stimulus, we show the ground-truth image, our reconstruction, and results from competing methods. 
Our approach consistently outperforms existing methods in the challenging zero-shot cross-subject setting.
}
\label{fig:results}
\end{figure*}

\subsection{Experimental setup}
\noindent
{\bf Implementation Details.} 
We first train the universal fMRI autoencoder for $250{,}000$ iterations with a batch size of 352 using 8 NVIDIA H100 GPUs. The autoencoder contains approximately 1.27B parameters. The CLS token length is set to 4.
Both the Transformer encoder and decoder have 32 layers, with an embedding dimension of 1280 and 16 attention heads. We optimize the model using AdamW with a learning rate of $1\times10^{-4}$, a warm-up period of 20{,}000 steps, and a weight decay of 0.01.
For compositional latent modeling, the universal autoencoder is frozen and only the Latent Factorization--Composition Module (LFCM) is trained.
LFCM contains approximately 195M parameters. Training is performed for 50{,}000 iterations with a batch size of 128 on 2 NVIDIA H200 GPUs. We use the AdamW optimizer, with a learning rate of $5\times10^{-5}$, 10{,}000 warm-up steps, and a weight decay of 0.01.  
We adopt image features extracted by IP-Adapter SDXL Plus as ground-truth visual targets. Accordingly, the stimulus-driven code has a dimensionality of $16 \times 2048$, while the stimulus-independent nuisance code is represented by a single $1 \times 2048$ vector. In the UniCortex-fMRI dataset, paired samples arise from both repeated trials of the same subject viewing the same stimulus, which facilitates disentanglement of trial-specific nuisance variability, and different subjects viewing the same stimulus, which promotes separation of subject-specific factors. For datasets where exact stimulus pairs are sparse, we additionally inject small Gaussian noise to simulate trial-wise variability. 
During inference, we set the Classifier-Free Guidance (CFG)~\cite{ho2022classifier} scale to 5.0.

\noindent
{\bf Evaluation Metrics.} 
We evaluate reconstruction quality from both low-level visual fidelity and high-level perceptual consistency. Low-level consistency is assessed using four commonly adopted metrics: pixelwise correlation, LPIPS~\cite{zhang2018perceptual}, AlexNet(2), and AlexNet(5).
Pixelwise correlation measures the average Pearson correlation between each reconstructed image and its corresponding ground-truth image at the pixel level.
LPIPS computes perceptual distance based on deep features and captures local texture and appearance discrepancies.
AlexNet(2) and AlexNet(5) denote two-way classification accuracy using features extracted from the second and fifth convolutional layers of AlexNet~\cite{krizhevsky2012imagenet}, respectively. 
High-level consistency is evaluated by comparing semantic representations extracted from pretrained visual encoders. Specifically, EfficientNet-B1 (EffNet-B)~\cite{tan2019efficientnet} and SwAV-ResNet50~\cite{caron2020unsupervised} are used to compute the average feature distance between reconstructions and ground-truth images, measuring alignment in learned representation spaces.
In addition, Inception~\cite{szegedy2016rethinking} and CLIP~\cite{radford2021learning} are employed to perform two-way classification based on their corresponding feature embeddings, assessing whether reconstructions preserve discriminative high-level visual content.

\subsection{Main Results}

\subsubsection{Competitors}
We compare our method against two representative open-source fMRI-to-image reconstruction frameworks, MindBridge~\cite{wang2024mindbridge} and MindEye2~\cite{scotti2024mindeye2}, as well as our prior conference version, NeuroPictor~\cite{neuropictor}. 
In their original formulations, MindBridge and MindEye2 are designed around subject-specific fMRI encoders, where voxel dimensionality varies across subjects. To enable cross-subject evaluation, we equip these methods with our unified cortical surface, while preserving their original model architectures and training objectives. 
All competing methods are retrained on our proposed UniCortex-fMRI dataset.

\subsubsection{Quantitative results}

\begin{table*}[!t]
\caption{Quantitative comparison of zero-shot cross-subject brain decoding.
Baseline methods report averaged results, while \textbf{OURS} reports
dataset-wise performance and the overall average.}
\label{tab:recon_eval}
\centering
\setlength{\tabcolsep}{5.5pt}
\begin{tabular}{lcccccccccc}
\toprule
\multirow{2}{*}{\textsc{Method}} 
& \multirow{2}{*}{\textsc{Data}}
& \multicolumn{4}{c}{Low-Level} 
& \multicolumn{4}{c}{High-Level} \\
\cmidrule(l){3-6} \cmidrule(l){7-10}
& 
& PixCorr $\uparrow$ 
& LPIPS $\downarrow$ 
& AlexNet(2) $\uparrow$ 
& AlexNet(5) $\uparrow$ 
& Inception $\uparrow$ 
& CLIP $\uparrow$ 
& EffNet-B $\downarrow$ 
& SwAV $\downarrow$ \\
\midrule
NeuroPictor~\cite{neuropictor} 
& Avg. 
& .017 & .731 & 54.7\% & 57.4\% & 56.9\% & 56.9\% & .952 & .622 \\
MindBridge~\cite{wang2024mindbridge} 
& Avg. 
& .050 & .751 & 58.6\% & 62.9\% & 60.2\% & 60.6\% & .954 & .626 \\
MindEye2~\cite{scotti2024mindeye2} 
& Avg. 
& .095 & .739 & 58.2\% & 59.8\% & 57.9\% & 56.5\% & .948 & .616 \\
\midrule
\multirow{5}{*}{\textbf{OURS}} 
& NSD 
& .158 & .655 & 82.9\% & 85.7\% & 77.1\% & 77.5\% & .854 & .500 \\
& HCP 
& .133 & .667 & 74.4\% & 76.2\% & 76.1\% & 77.0\% & .771 & .443 \\
& BOLD5000 
& .083 & .704 & 67.3\% & 72.7\% & 67.1\% & 68.3\% & .920 & .594 \\
& NOD 
& .044 & .734 & 61.4\% & 69.0\% & 59.4\% & 61.8\% & .974 & .638 \\
\cmidrule(l){2-10}
\rowcolor{gray!10}
& \textbf{Avg.} 
& \textbf{.104} & \textbf{.690} & \textbf{71.5\%} & \textbf{75.9\%}
& \textbf{69.9\%} & \textbf{71.2\%} & \textbf{.880} & \textbf{.544} \\
\bottomrule
\end{tabular}
\end{table*}

\begin{figure*}[!t]
\centering
\includegraphics[width=\linewidth]{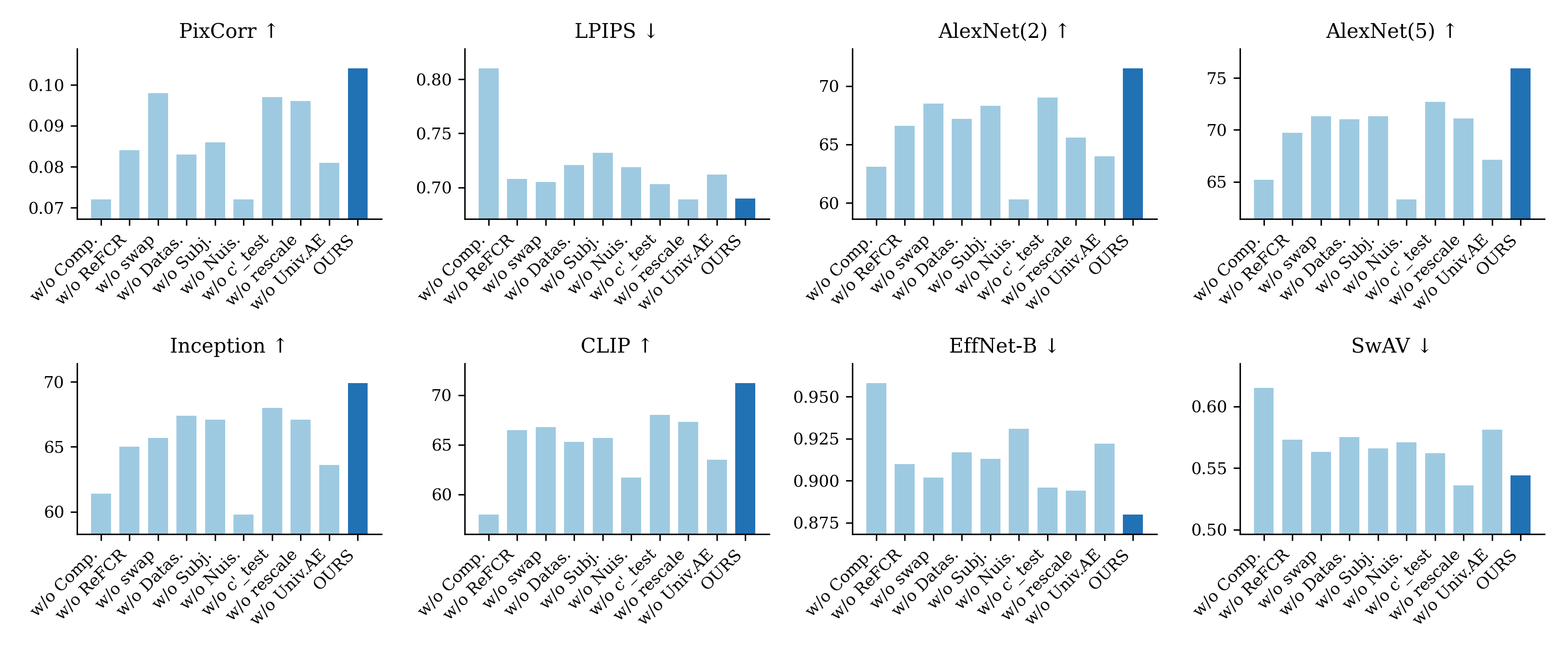}
\caption{Ablation study on key components of our framework. Each subfigure corresponds to one evaluation metric. Light blue bars indicate results from ablation experiments, while dark blue bars show the full model. 
}
\label{fig:ablation}
\end{figure*}

Table~\ref{tab:recon_eval} reports the quantitative comparison between our method and existing fMRI-to-image reconstruction frameworks under the UniCortex-fMRI dataset. AVG denotes the standard UniCortex setting obtained by averaging metrics across the four datasets. Overall, our approach consistently and substantially outperforms all competing methods across both low-level and high-level metrics. Compared with NeuroPictor, MindBridge, and MindEye2, we observe significant gains in perceptual similarity and feature-based recognition scores, indicating that our reconstructions better preserve visual content under cross-subject evaluation.
A key limitation of previous methods, including NeuroPictor, MindBridge, and MindEye2, is their lack of explicit modeling of dataset-, subject- and trial-specific factors. In addition, NeuroPictor does not provide a unified visual reference across subjects. As a result, these approaches struggle to generalize to unseen subjects, even when retrained on the unified cortical surface. In contrast, our method explicitly separates stimulus-driven content from subject- and trial-specific variations via compositional latent modeling, enabling robust cross-subject generalization and more accurate reconstructions.
Examining dataset-wise results, NSD achieves the highest overall scores among all datasets, likely benefiting from its higher signal-to-noise ratio and dense stimulus repetitions during acquisition. Performance on HCP and BOLD5000 remains competitive, while NOD shows relatively lower scores, reflecting increased variability in the dataset.
These results demonstrate that our compositional latent framework effectively captures stable stimulus representations and translates them into faithful visual reconstructions, validating its effectiveness for zero-shot cross-subject fMRI-to-image decoding.

\subsubsection{Qualitative results}

Figure~\ref{fig:results} visualizes fMRI-to-image reconstructions for unseen subjects in a zero-shot inference setting, comparing our method with representative baselines, including NeuroPictor~\cite{neuropictor}, MindBridge~\cite{wang2024mindbridge}, and MindEye2~\cite{scotti2024mindeye2}. 
In this challenging scenario, our approach reliably reconstructs both the semantic content and the spatial structure of the original stimuli. 
In contrast, competing methods often produce blurred or semantically inconsistent reconstructions, reflecting their limited ability to generalize to unseen subjects. 
These failures can be attributed to the lack of explicit disentanglement between stimulus-driven content and subject- or trial-specific variations in previous approaches. 
By modeling these factors separately, our method generates robust and faithful reconstructions, maintaining high perceptual fidelity across multiple datasets. This demonstrates the effectiveness of our compositional latent framework for zero-shot cross-subject fMRI decoding.

\subsection{Ablation Study}

\begin{figure*}[!t]
\centering
\includegraphics[width=\linewidth]{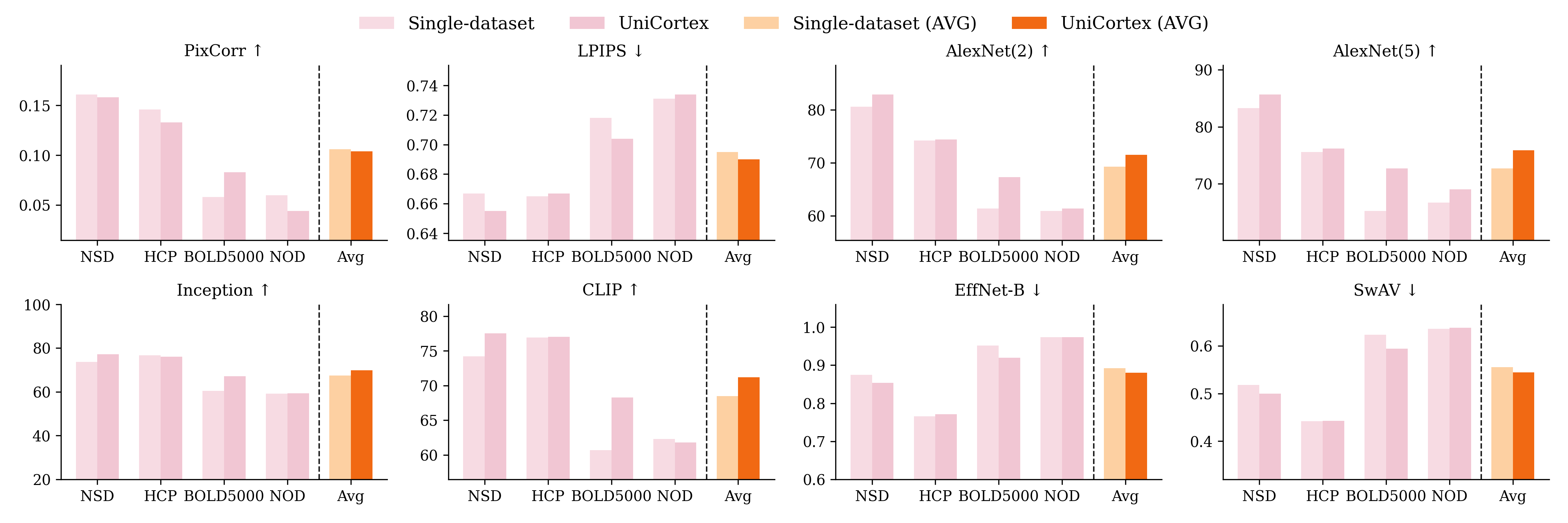}
\caption{Comparison of multi-dataset integration versus single-dataset training. Each panel shows one evaluation metric. Multi-dataset pretraining consistently improves most metrics, demonstrating more stable and generalizable stimulus representations.}
\label{fig:ablation_dataset}
\end{figure*}

We conduct a comprehensive ablation study to evaluate the contributions of the Latent Factorization–Composition Module (LFCM), compositional components, the universal fMRI autoencoder, inference-stage refinements, and multi-dataset integration. Figures~\ref{fig:ablation} and~\ref{fig:ablation_dataset} provide visual summaries of these effects across eight evaluation metrics, illustrating the performance drop when a specific component is removed or a design choice is altered. These results highlight the critical role of each component in achieving robust zero-shot cross-subject fMRI-to-image reconstruction.

\subsubsection{Latent Factorization–Composition Module}
Removing the Compositor (\textit{w/o Comp.}) reduces the model to a direct mapping from fMRI latent to stimulus code, losing the ability to disentangle stimulus-driven content from nuisance factors. This significantly degrades performance because the model can no longer isolate subject- and trial-independent representations. Removing the Re-Factorizing Consistency Regularization (\textit{w/o ReFCR}) weakens the factorizer's robustness to off-manifold latents, leading to less stable reconstructions. Similarly, disabling the pairwise swapping of stimuli across subjects and trials (\textit{w/o swap}) limits the exposure of the factorizer to diverse surrogate latents, reducing its ability to generalize to unseen subjects.

\subsubsection{Compositional Components}
We further evaluate the contribution of each compositional component by removing the dataset condition (\textit{w/o Datas.}), subject condition (\textit{w/o Subj.}), or nuisance component (\textit{w/o Nuis.}).
Removing any of these components consistently degrades performance, indicating that omitting any factor leads to an incomplete modeling of fMRI variability.
In particular, removing the nuisance component results in the most severe performance drop.
Without a nuisance term, the model is forced to explain trial-specific and non-stimulus-related variability directly through the stimulus-driven code, coupling semantic content with residual noise.
This introduces ambiguity in the latent space and makes the stimulus code less identifiable.

\subsubsection{Universal Autoencoder}
Training without a pre-trained universal autoencoder (\textit{w/o UnivAE}) also reduces performance. Although the UniCortex-fMRI dataset contains over 800k fMRI samples across 219 subjects, it lacks the scale of large pretraining datasets like UKB needed to capture a high-capacity shared latent space. The universal autoencoder provides a robust latent foundation that enhances subsequent factorization and decoding.

\subsubsection{Inference-Stage Refinements}
We examine the contributions of our inference-stage refinements. Removing surrogate-based re-factorizing (\textit{w/o $\mathbf{c}_{\text{te}}^{\prime}$}) forces direct use of the initial estimated latent, resulting in decreased robustness to subject-specific distributional shifts. Disabling rescaling for the predicted stimulus-driven latents (\textit{w/o rescale}) also degrades performance, as the predicted latent may deviate in scale from the training feature distribution, reducing stability for the diffusion model.

\subsubsection{Multi-Dataset Integration}
To assess the effect of combining multiple datasets, we compare single-dataset training (\textit{Single dataset}) with multi-dataset joint training (\textit{UniCortex}). Figure~\ref{fig:ablation_dataset} visualizes the results across eight evaluation metrics. Overall, integrating all four datasets improves performance for the majority of metrics. In particular, seven out of eight metrics show gains under multi-dataset training, with especially notable improvements in AlexNet and CLIP classification accuracies. This demonstrates that multi-dataset pretraining yields more stable and generalizable stimulus representations than training on individual datasets alone.  
PixCorr shows a slightly lower value for multi-dataset training compared to some individually trained datasets. This likely arises because single-dataset training allows the model to more closely fit the dataset-specific image distribution, producing reconstructions that are pixelwise closer to each dataset’s stimuli. In contrast, multi-dataset training encourages learning a more generalized representation that balances across diverse datasets.  
Among the four datasets, NSD and BOLD5000 benefit most from multi-dataset integration. Both datasets feature a wide range of stimulus images (thousands per subject) but relatively few subjects (8 for NSD, 4 for BOLD5000). Introducing additional subjects from other datasets improves the model’s ability to disentangle subject-specific factors and strengthens cross-subject generalization, leading to larger performance gains.  

These ablation results collectively confirm that each component 
contributes critically to the robust and accurate reconstruction of visual stimuli from fMRI, particularly under the challenging zero-shot cross-subject setting.

\begin{figure}
\centering
\includegraphics[width=\linewidth]{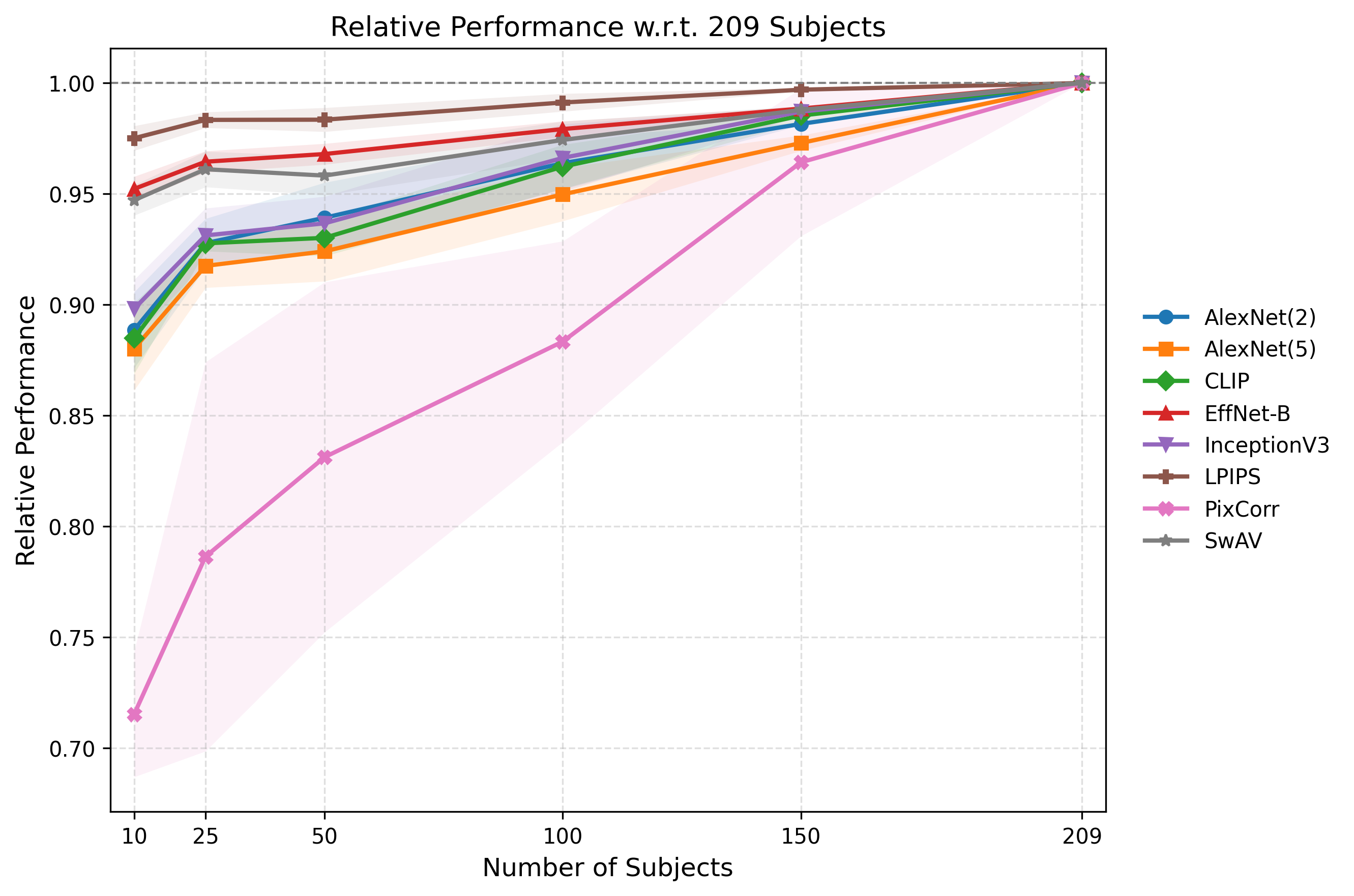}
\caption{Impact of subject scale on cross-subject generalization. 
We show the relative performance of zero-shot fMRI-to-image reconstruction for different numbers of subjects, normalized by the full 209-subject performance (set to 1.0). Lines and points represent the mean across five random samplings; shaded regions indicate standard deviation. 
}
\label{fig:subnum_metrics_relative}
\end{figure}

\subsection{Impact of Subject Scale}
We investigate how the number of subjects in the training data affects zero-shot cross-subject fMRI-to-image reconstruction performance, as shown in Figure~\ref{fig:subnum_metrics_relative}. We consider subsets of 10, 25, 50, 100, and 150 subjects, and normalize all metrics relative to the full 209-subject performance (set to 1.0). For each subset size, we perform five random samplings of subjects. Lines and points represent the mean across these samplings, and the shaded regions indicate standard deviation.  
Subjects are sampled in a stratified manner across datasets. Ensuring that each subset contains at least one subject from each dataset, the remaining subjects are sampled proportionally to dataset sizes.  
The results demonstrate that performance consistently improves with increasing subject count across all metrics. The steepest improvement occurs between 10 and 25 subjects, reflecting that small datasets lack sufficient inter-subject variability to learn robust, generalizable representations. As the number of subjects approaches 150 and 209, the performance curve flattens, indicating diminishing returns when subject diversity is nearly saturated. These findings highlight the importance of including a sufficient number of subjects to capture stable, cross-subject stimulus representations for zero-shot fMRI decoding.

\section{Conclusion}

In this work, we explored the challenge of cross-subject fMRI-to-image reconstruction in zero-shot scenario, a paradigm complicated by substantial inter- and intra-subject variability in cortical responses. To enable principled evaluation and learning, we first constructed \textbf{UniCortex-fMRI} by integrating multiple heterogeneous visual-stimulus datasets and providing broad coverage of subjects and stimuli.  
Building on this resource, we proposed \textbf{PictorialCortex}, a compositional latent framework that explicitly disentangles stimulus-driven visual information from dataset-, subject-, and trial-specific nuisance factors. Leveraging a universal cortical latent space pretrained on large-scale fMRI data and a Latent Factorization–Composition Module with paired factorization and re-factorizing consistency regularization, our method achieves robust zero-shot reconstruction for previously unseen subjects.  
Extensive experiments demonstrate that PictorialCortex faithfully recovers semantic and spatial content across diverse datasets, benefits from multi-dataset pretraining, and scales effectively with the number of subjects. Overall, our work provides both a principled dataset and a scalable, generalizable decoding framework, supporting further study of subject-agnostic neural representations and cross-subject visual reconstruction.

\section*{Acknowledgments}
We thank Yun Wang for valuable discussions on neuroscience that helped inform this work.


\bibliographystyle{IEEEtran}
\bibliography{main}

% \newpage

% ---- Supplementary Material ----
\clearpage
\noindent \textbf{\Large Supplementary Material}
\vspace{0.1in}
\setcounter{section}{0}

\section{Implementation Details}

\noindent \textbf{Universal fMRI Autoencoder.}
We first train the universal fMRI autoencoder to obtain a shared cortical latent space that can represent fMRI signals across multiple subjects. 
The autoencoder is trained on the UKB dataset \cite{miller2016multimodal} for $250{,}000$ iterations with a batch size of 352 on 8 NVIDIA H100 GPUs. 
It contains approximately 1.27B parameters in total. 
The CLS token length is set to 4 to encode global cortical information. 
Both the Transformer encoder and decoder consist of 32 layers, with an embedding dimension of 1280 and 16 attention heads. 
We optimize the autoencoder using AdamW with a learning rate of $1\times10^{-4}$, a warm-up period of 20{,}000 steps, and a weight decay of 0.01.

\noindent \textbf{Latent Factorization--Composition Module (LFCM).}
After training the universal autoencoder, its parameters are frozen to preserve the learned universal latent space.
We then train the Latent Factorization--Composition Module (LFCM) for compositional latent modeling.
LFCM contains approximately 195M parameters and is trained for 50{,}000 iterations with a batch size of 128 on 2 NVIDIA H200 GPUs.
We use AdamW with a learning rate of $5\times10^{-5}$, 10{,}000 warm-up steps, and a weight decay of 0.01.

\noindent \textbf{Latent representations and visual targets.}
We adopt image features extracted by IP-Adapter SDXL Plus~\cite{ipadapter} as ground-truth visual targets to guide the learning of stimulus-driven codes.
Accordingly, the stimulus-driven latent code $\mathbf{c}$ has a dimensionality of $16 \times 2048$,
while the stimulus-independent nuisance code $\mathbf{n}$ is represented by a single $1 \times 2048$ vector.
We adopt image features extracted by IP-Adapter SDXL Plus~\cite{ipadapter} as ground-truth visual targets
to guide the learning of stimulus-driven codes.
Accordingly, the stimulus-driven latent code $\mathbf{c}$ has a dimensionality of $16 \times 2048$,
while the stimulus-independent nuisance code $\mathbf{n}$ is represented by a single $1 \times 2048$ vector.
Both subject embeddings $e^{\mathrm{sub}}$ and dataset embeddings $e^{\mathrm{data}}$
are implemented as learnable $1 \times 2048$ vectors, which are incorporated into the latent inputs
to provide contextual conditioning for both the Factorizer and the Compositor.

\noindent \textbf{Architecture of the Factorizer and Compositor.}
Both the Factorizer and the Compositor are implemented as symmetric Transformer blocks,
each consisting of two Transformer layers with an embedding dimension of 2048 and 32 attention heads.
In the Factorizer, a set of 17 learnable queries of dimension $17\times2048$ is used to extract the latent components.
The first 16 queries produce the stimulus-driven code $\mathbf{c}$ ($16\times2048$) corresponding to the visual target features extracted by IP-Adapter,
while the last query produces the nuisance code $\mathbf{n}$ ($1\times2048$).
In the Compositor, 4 learnable queries of dimension $4\times2048$ are employed to reconstruct the universal latent,
corresponding to the 4 CLS tokens of the autoencoder latent.
Before being fed into the Transformer blocks, the learnable subject and dataset embeddings are added to the latent inputs as conditioning signals.
The Compositor queries attend to the concatenated stimulus-driven and nuisance codes to synthesize surrogate universal latents, 
while the Factorizer queries attend to the input universal latent to decompose it into the corresponding components.

\begin{figure*}
\centering
\includegraphics[width=\linewidth]{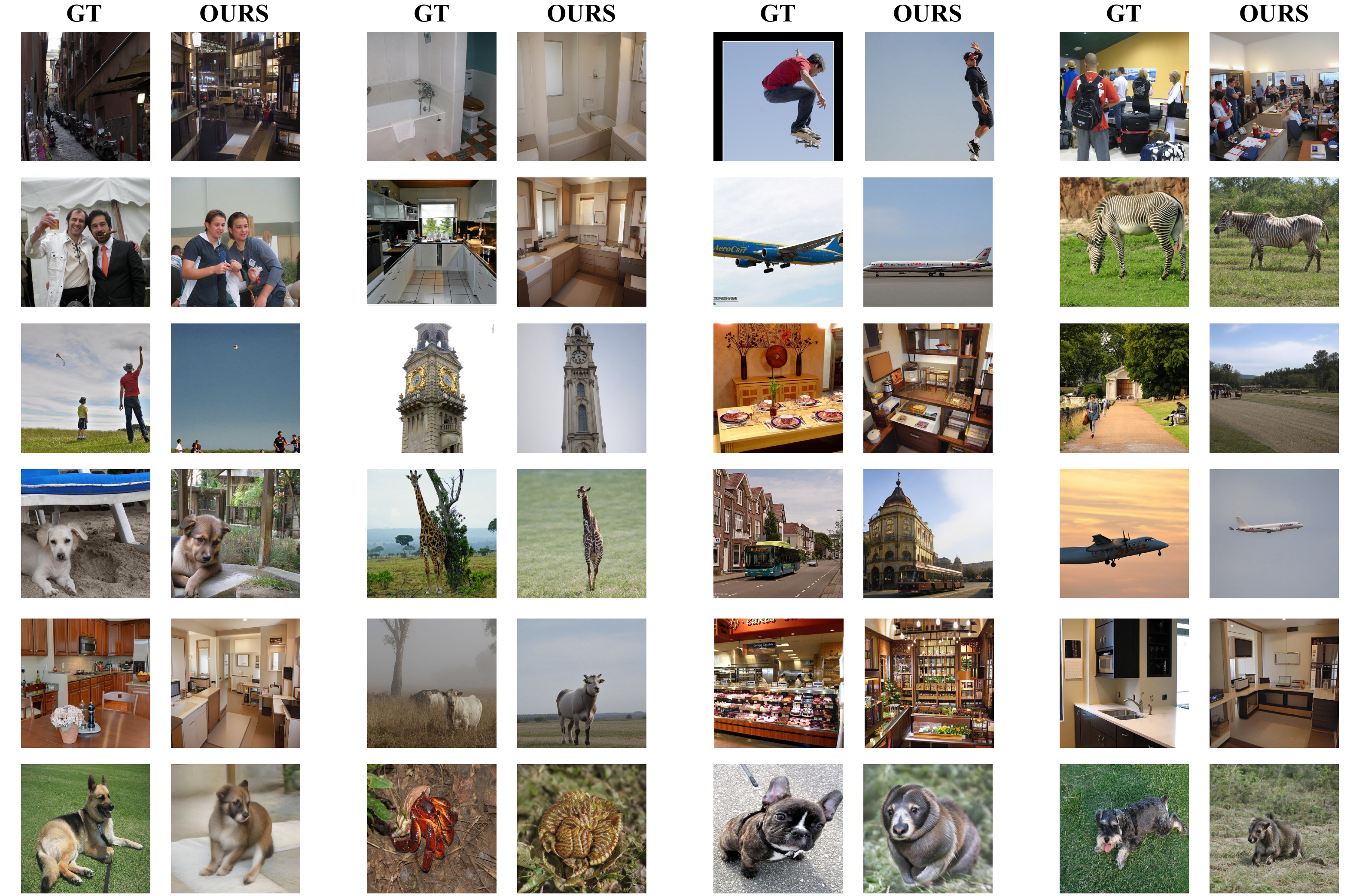}
\caption{Additional visualizations of zero-shot cross-subject fMRI-to-image reconstruction.
}
\label{fig:supp_vis}
\end{figure*}

\begin{figure*}
\centering
\includegraphics[width=\linewidth]{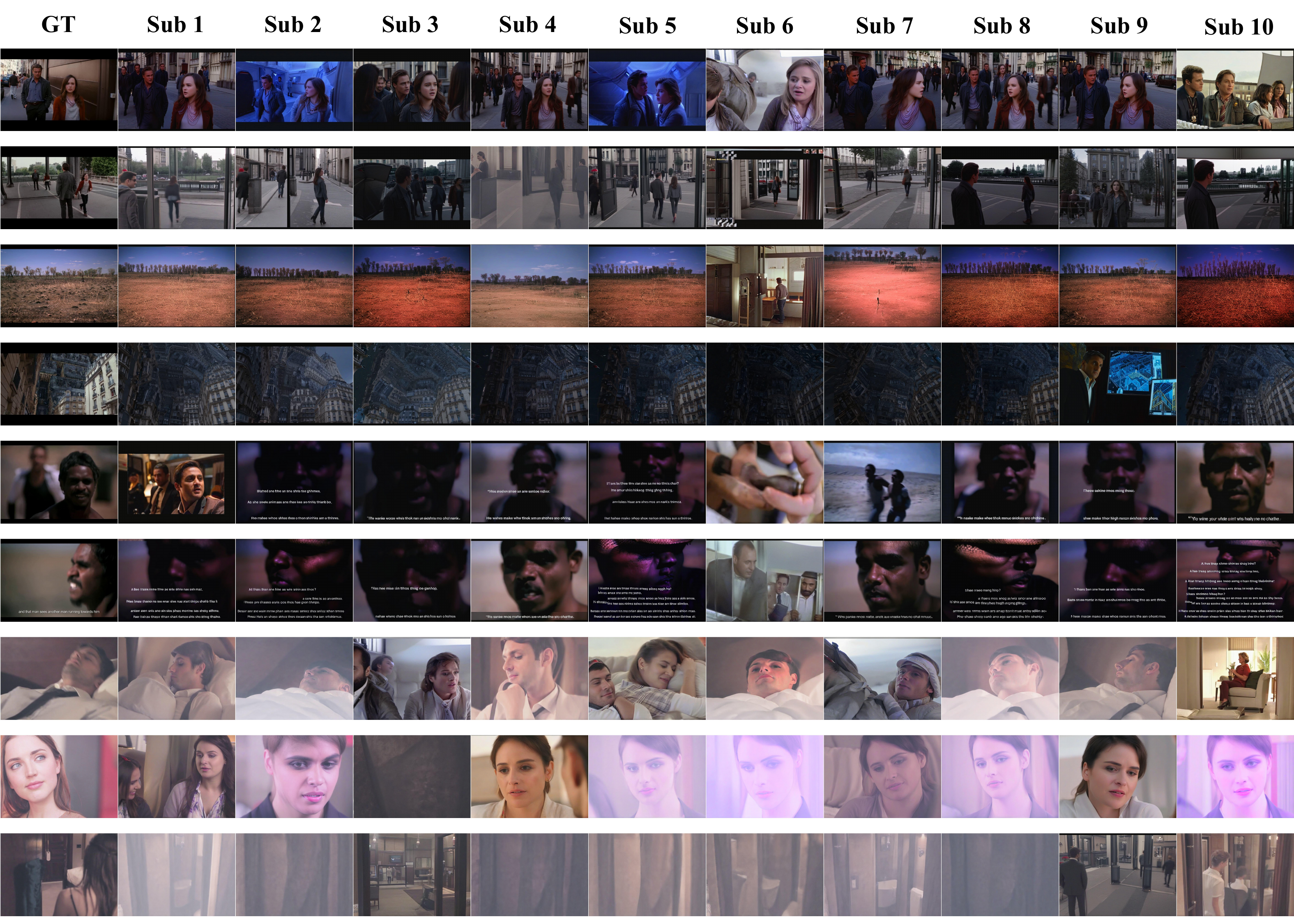}
\caption{
Per-subject zero-shot cross-subject fMRI-to-image reconstruction results on unseen subjects from the HCP-Movie dataset.
The first column shows the ground-truth image, while columns 2--11 present reconstructions for subjects 1--10, respectively.
}
\label{fig:supp_hcp}
\end{figure*}

\begin{figure*}
\centering
\includegraphics[width=\linewidth]{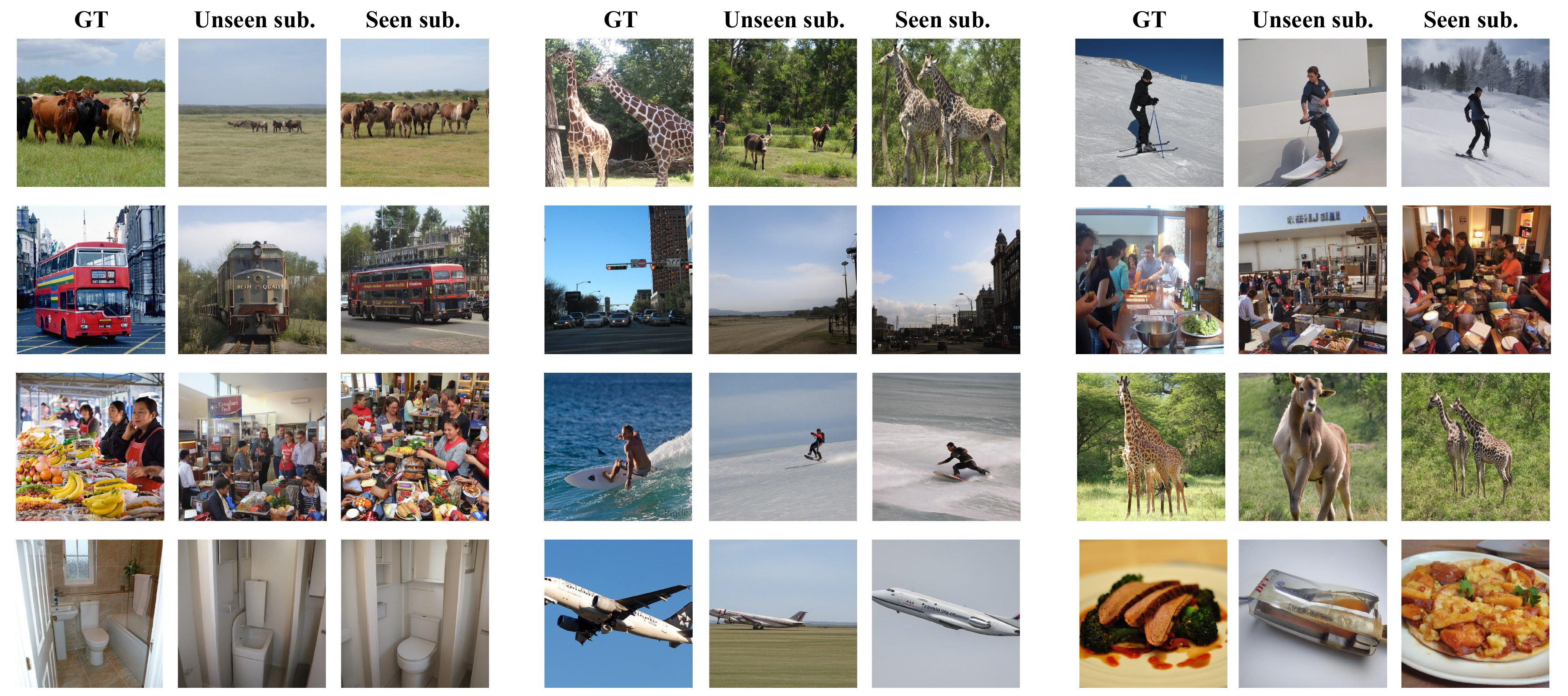}
\caption{Visually Comparison on the reconstructed images from unseen and seen subjects by our model. 
In each group, the first column shows the ground-truth image,
the second column presents the reconstruction from an unseen subject,
and the third column shows the reconstruction from a seen subject.
}
\label{fig:supp_seen}
\end{figure*}

\begin{figure*}
\centering
\includegraphics[width=\linewidth]{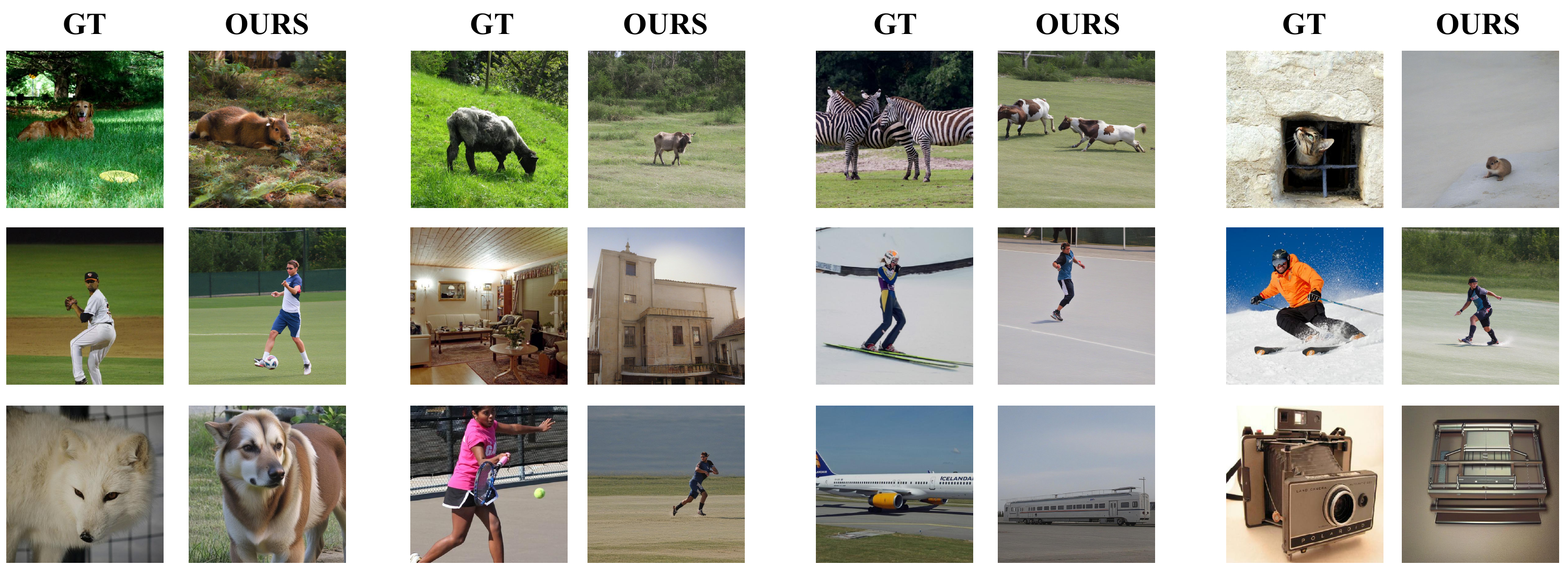}
\caption{Visualizations for some failure cases.
}
\label{fig:supp_fail}
\end{figure*}

\begin{figure*}[!h]
\centering
\includegraphics[width=\linewidth]{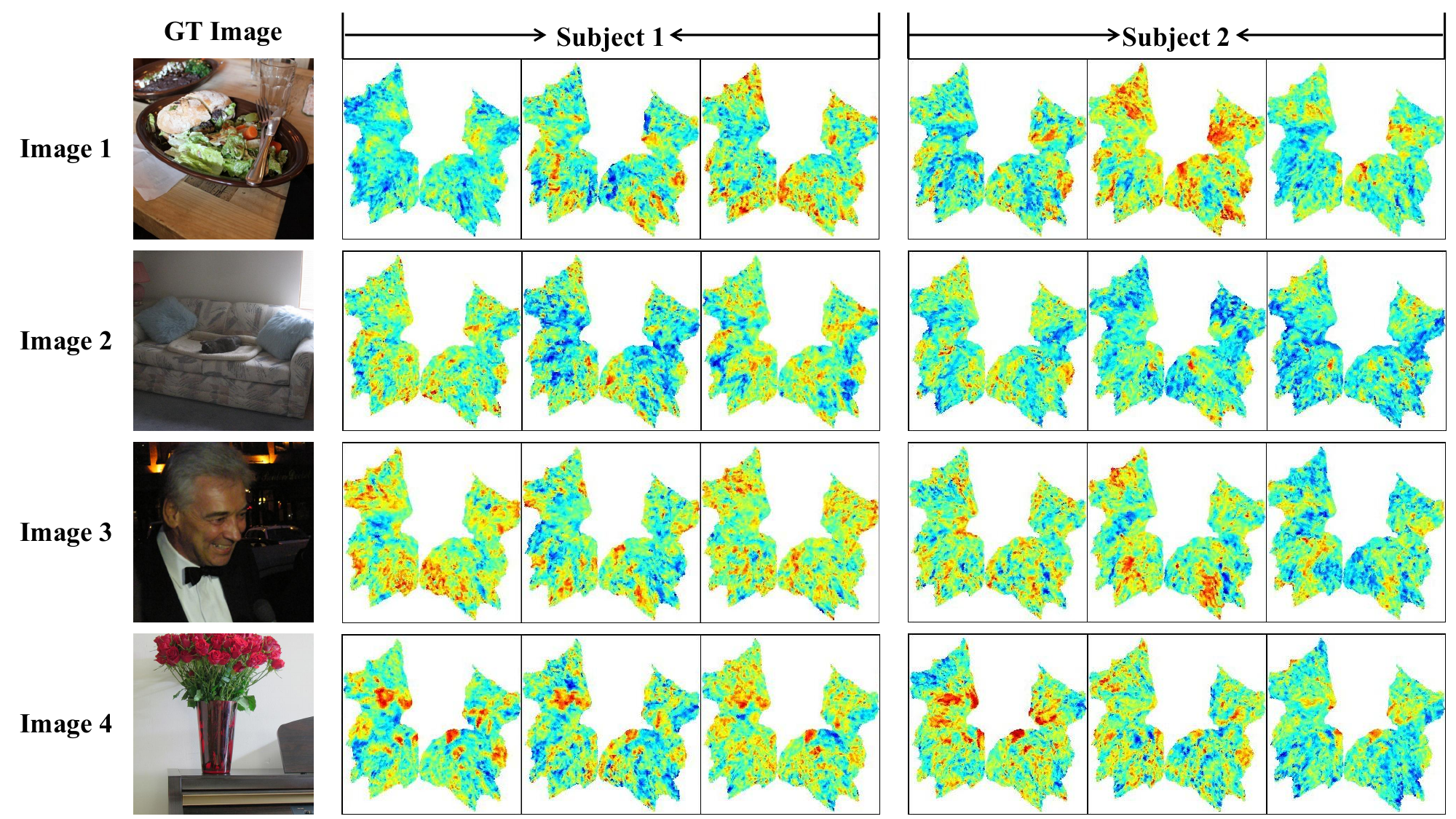}
\caption{
Visualizations of fMRI responses from two subjects viewing the same visual stimuli.
In each row, the first column shows the ground-truth image.
Columns 2--4 display three fMRI responses from Subject~1 under repeated presentations of the same image
, while columns 5--7 show the corresponding three fMRI responses from Subject~2.
}
\label{fig:supp_fmri}
\end{figure*}

\section{More Visualizations}

We provide additional visualizations of our method in zero-shot cross-subject fMRI-to-image reconstruction.
Figure~\ref{fig:supp_vis} presents reconstruction results across multiple datasets.
All results are obtained from unseen subjects that are excluded from training.
Specifically, rows 1--4 correspond to the NSD dataset, row 5 shows results from BOLD5000, and row 6 corresponds to NOD.
Figure~\ref{fig:supp_hcp} further reports per-subject reconstruction results on the HCP-Movie dataset.
Subjects 1--10 are treated as unseen during training and evaluated individually at test time.
The visualizations demonstrate that our approach maintains stable reconstruction quality across different unseen subjects.

We additionally provide a comparison between reconstructions from unseen subjects (i.e., zero-shot cross-subject reconstruction in our setting) and seen subjects used during training, as shown in Figure~\ref{fig:supp_seen}.
In each group, the first column shows the ground-truth image,
the second column corresponds to the reconstruction from an unseen subject,
and the third column shows the reconstruction from a seen subject.
All reconstructions are generated from test-set stimuli that are never observed during training.
Although our method demonstrates robust zero-shot generalization to unseen subjects,
reconstructions from unseen subjects are generally slightly less accurate than those from seen subjects,
which is expected due to unavoidable subject-specific domain shifts.
Reconstructions from seen subjects typically exhibit more precise semantic details and stronger spatial consistency.
These results suggest that further improving generalization to unseen subjects remains an important and promising direction for future research.

Figure~\ref{fig:supp_fmri} visualizes fMRI responses from different subjects and repeated trials under identical visual stimuli.
For each image, we show multiple fMRI responses recorded from two subjects across repeated presentations.
In each row, the first column shows the ground-truth image.
Columns 2--4 display three fMRI responses from Subject~1 under repeated presentations of the same image
, while columns 5--7 show the corresponding three fMRI responses from Subject~2.
As can be observed, even when viewing the same image, the elicited fMRI responses vary across subjects, and also across trials within the same subject.
This visualization shows the pronounced inter-subject and intra-subject variability inherent in fMRI signals.

\section{Failure Cases}

Figure~\ref{fig:supp_fail} illustrates representative failure cases of our method.
Despite our advancements, zero-shot cross-subject fMRI-to-image reconstruction remains challenging, particularly when discriminating semantically similar concepts.
As shown in the Figure~\ref{fig:supp_fail}, some reconstructions preserve coarse spatial layouts but drift toward semantically related yet incorrect concepts, such as confusing a crouching dog with a squirrel, a white airplane with a white train, zebras with cows on grasslands, or foxes with dogs.
For scenes involving humans, the model often captures approximate poses or spatial configurations but sometimes misinterpreting or omitting some semantic attributes, e.g., confusing skiing with surfing, or generating a running person without the associated object such as a tennis racket.
These failure cases indicate that, for unseen subjects, fine-grained semantic generalization remains difficult under cross-subject cortical variability.

\section{Broader Impact}

Beyond model performance, this work makes broader contributions at both the dataset and methodological levels.

First, we provide a curated dataset that enables systematic evaluation of zero-shot cross-subject fMRI decoding by using heterogeneous datasets. Unlike prior studies that focus on narrowly controlled settings or subject-specific training, our dataset and evaluation protocol explicitly target cross-subject generalization without subject-specific finetuning. This establishes a unified testbed for comparing representation learning methods under realistic and challenging transfer scenarios. 

Second, by organizing heterogeneous fMRI data into a unified representational and evaluation framework, our work offers a scalable pathway for future research. As neuroimaging studies continue to grow in scale and diversity, the ability to aggregate data across datasets, experimental protocols, and populations becomes increasingly important. Our dataset construction and evaluation strategy provide a practical foundation for scaling cross-subject neurodecoding research beyond isolated datasets, encouraging more reproducible and comparable progress in the field.

Third, at the methodological level, PictorialCortex introduces a paradigm of interpretable, compositional latent formulation for jointly modeling fMRI activation patterns. By explicitly modeling cortical responses using stimulus-driven, subject-related, dataset-related, and nuisance components within a shared latent space, our approach provides a structured lens for studying how different sources of variability interact in neural signals.

Looking forward, this modeling philosophy may inspire future work on neural representation analysis, cross-population neuroscience, and brain–machine interfaces, particularly in settings that demand 
generalization. We hope that the combination of scalable data resources and compositional latent modeling will foster more robust, generalizable neurodecoding systems.

\section{Limitations and Future work}
Although PictorialCortex demonstrates strong performance, several limitations remain.
The framework relies on a pretrained autoencoder to establish the universal cortical latent space. While this provides stability and scalability, errors or biases introduced at this stage may propagate to downstream compositional modeling. Joint or adaptive optimization of representation learning and factor disentanglement remains an open direction.
In addition, the model focuses on static visual perception. Extending the compositional latent framework to temporal dynamics, such as video perception or mental imagery over time, represents an important and nontrivial future direction.
Overall, we view PictorialCortex as a step toward cross-subject neural decoding, and anticipate that future work combining richer experimental designs and more expressive latent models will further advance this line of research.

\end{document}